\documentclass{article}
\usepackage{kr}

\usepackage{times}
\usepackage{soul}
\usepackage{url}
\usepackage[hidelinks]{hyperref}
\usepackage[utf8]{inputenc}
\usepackage[small]{caption}
\usepackage{latexsym, amsmath, amssymb, amsfonts, mathrsfs}
\usepackage{xspace}
\usepackage[T1]{fontenc}
\usepackage{float}
\usepackage{wrapfig}
\usepackage{color}
\usepackage{url}
\usepackage[pdftex]{graphicx}
\usepackage{enumitem}
\usepackage[ruled,linesnumbered,noresetcount,vlined]{algorithm2e}
\usepackage{todonotes}
\usepackage{picinpar}
\usepackage{multirow}
\usepackage{subcaption}
\usepackage[normalem]{ulem}
\usepackage{xcolor}
\usepackage{booktabs}

\usepackage[ruled, vlined, linesnumbered]{algorithm2e}
\usepackage{comment}
\usepackage{makecell}
\usepackage{float}
\usepackage{array}
\usepackage{placeins}

\allowdisplaybreaks

\newcommand{\fml}[1]{\mathcal{#1}}

\newcommand{\KB}{\mathrm{KB}}

\newcommand{\beitemize}{\begin{list}{$\bullet$}{\topsep=1.5pt \parsep=0pt \itemsep=1pt \leftmargin=1em }} 
\newcommand{\enitemize}{\end{list}}

\newcommand{\beenumerate}{\hspace{-0.5in} \begin{enumerate}\topsep=1pt \parsep=0pt \itemsep=-3pt} \newcommand{\enenumerate}{\end{enumerate}}

\newcommand{\belist}{\begin{list}{$\bullet$}{\topsep=1.5pt \parsep=0.5pt \itemsep=1pt \leftmargin=2.25em \labelwidth=1.0em \labelsep=0.5em \partopsep=1.5pt}} 
\newcommand{\enlist}{\end{list}}

\newtheorem{theorem}{{\bf Theorem}}

\newtheorem{definition}{{\bf Definition}}

\newtheorem{example}{{\bf Example}}

\newenvironment{sproof}{\noindent {\sc Proof. }}{\hfill$\Box$}

\newboolean{includeMemo}
\setboolean{includeMemo}{true} 

\newcommand{\memoside}[1]{\ifthenelse{\boolean{includeMemo}}{\todo[caption={},color=green!20!]{{\footnotesize #1}}}}
\newcommand{\memo}[1]{\ifthenelse{\boolean{includeMemo}}{\todo[inline,caption={},color=green!20!]{#1}}}
\newcommand{\memob}[1]{\ifthenelse{\boolean{includeMemo}}{\todo[inline,caption={},color=blue!20!]{#1}}}

\newcommand{\ignore}[1]{}

\newcommand{\squishlist}{
\begin{list}{{{\small{$\bullet$}}}}
{\setlength{\itemsep}{3pt}      
\setlength{\parsep}{3pt}
\setlength{\topsep}{3pt}       
\setlength{\partopsep}{3pt}
\setlength{\leftmargin}{1em} 
\setlength{\labelwidth}{1em}
\setlength{\labelsep}{0.5em} } }
\newcommand{\squishend}{  \end{list}}

\newcommand{\squishenum}{
\begin{list}{$\bullet$}{ 
    \setlength{\itemsep}{1pt}
    \setlength{\parsep}{0pt}
    \setlength{\topsep}{1.5pt}
    \setlength{\partopsep}{0pt}
    \setlength{\leftmargin}{2em}
    \setlength{\labelwidth}{1.5em}
    \setlength{\labelsep}{0.5em} } }

\newcommand{\citet}[1]{\citeauthor{#1}~\citeyear{#1}}
\newcommand{\citep}{\cite}

\usepackage{ulem}
\renewcommand{\underline}{\uline}

\title{Dialectical Reconciliation via Structured Argumentative Dialogues\footnote{This paper has been published in KR 2024.}}

\author{Stylianos Loukas Vasileiou$^1$\and
Ashwin Kumar$^1$\and
William Yeoh$^1$\and
Tran Cao Son$^2$\and
Francesca Toni$^3$ \\
\affiliations
$^1$Washington University in St. Louis\\
$^2$New Mexico State University\\
$^3$Imperial College London\\
\emails
\{v.stylianos, ashwinkumar, wyeoh\}@wustl.edu,
stran@nmsu.edu,
f.toni@imperial.ac.uk}

\begin{document}

\maketitle

\begin{abstract}

We present a novel framework designed to extend \textit{model reconciliation} approaches, commonly used in human-aware planning, for enhanced human-AI interaction. By adopting a structured argumentation-based dialogue paradigm, our framework enables \textit{dialectical reconciliation} to address knowledge discrepancies between an \textit{explainer} (AI agent) and an \textit{explainee} (human user), where the goal is for the \textit{explainee to understand the explainer's decision}. We formally describe the operational semantics of our proposed framework, providing theoretical guarantees. We then evaluate the framework's efficacy ``in the wild'' via computational and human-subject experiments. Our findings suggest that our framework offers a promising direction for fostering effective human-AI interactions in domains where explainability is important. 

\end{abstract}

\section{Introduction}

The rapid advancement and integration of AI 
into various aspects of daily life underscore the need for systems that are not only effective and adaptable but also explainable and understandable to human users. In response, within the subfield of \textit{human-aware planning} (HAP)~\citep{kambhampati2019synthesizing}, researchers focus on developing AI agents capable of explaining their decisions and actions in a manner comprehensible to human users. At the heart of HAP is the concept of the \textit{model reconciliation process} (MRP)~\citep{chakraborti2017plan}, 
aimed at aligning the models of an AI agent and a human user when those models
diverge in a way that a decision generated from 
the former is inexplicable in the 
latter. The reconciliation process essentially involves generating an \textit{explanation} from the AI agent's model such that when it is used to update the human user model, the AI agent's decision becomes explicable.~Although MRP originated to address planning problems ~\cite{sreedharan2021foundations}, it has been extended to problems beyond planning that admit 
logic-based representations ~\cite{son2021model,vasileioulogic}

However, most MRP approaches face two significant challenges. First, they often assume that the AI agent has access to an a-priori human user model. This assumption can lead to misunderstandings, as the agent might base its explanations on an inaccurate or incomplete understanding of the human user's knowledge of the underlying task. Second, they typically rely on single-shot interactions. While this may be sufficient when the user needs to quickly understand a decision or when the underlying task is relatively simple, it may fail to work for more complex decisions and tasks that require a deeper understanding from the human user, especially when there is substantial knowledge discrepancy between the AI agent and human user models.

These limitations give rise to the following pressing question: ``\textit{How can we effectively help the human user understand the AI agent's decisions}?'' Looking at the literature on cognitive science and psychology, we find some inspiration -- people learn and understand better when they engage in argumentation-based dialogues. Such dialogues engage the participants' cognitive abilities, enhancing learning and understanding through active engagement, reconstruction, and assimilation of information \cite{mercier2011humans}. In other words, reconciliation is \textit{dialectical}.

Motivated by this, in this paper we propose \textit{Dialectical Reconciliation via Structured Argumentative Dialogues} (DR-Arg), a novel framework wherein an \textit{explainer} (AI agent) and an \textit{explainee} (human user) engage in a \textit{dialectical reconciliation dialogue} aimed at helping the explainee understand the explainer's decisions. DR-Arg does not rely on predefined human user models but instead allows for a dynamic interaction that facilitates a more nuanced exchange of information. Importantly, the goal of DR-Arg is to \textit{enhance the explainee's understanding of the explainer's decisions}, even if the explainee ultimately disagrees with those decisions. This sets our framework apart from traditional argumentation frameworks that aim to achieve mutual agreement through persuasion \citep{gordon1994,prakken2006formal,parsons2003properties}.

From a technical standpoint, our work builds upon and extends previous efforts in argumentation-based dialogues \citep{black2021argumentation} and is formalized using a game-theoretic approach to dialogues \citep{hamblin1970fallacies,hamblin1971mathematical}. We formally define the notion of a dialectical reconciliation dialogue, describe its operational semantics with the use of \emph{structured (deductive) argumentation} \citep{besnard2001logic}, and provide theoretical guarantees regarding \textit{termination} and \textit{success}. 
Then, we turn to evaluating our framework ``in the wild''.
First, 
we discuss the concept of \textit{explainee understanding} in the context of these interactions and present a simple method for approximating it. Finally, we empirically evaluate the effectiveness of DR-Arg in both computational and human-user experiments, demonstrating its efficacy and potential for enhancing human-AI interactions in the real world.

\subsection{Motivating Example}
\label{sec:motivating-example}

To illustrate the potential of our approach, consider a scenario where a human user, Alice, is tasked with troubleshooting an AI home assistant robot, named ``Roomie'', that appears to be disconnected from the internet. Alice is provided with a set of prompts to help diagnose the problem, such as checking the associated mobile app and verifying Roomie's connection to the internet via a wired connector.

Initially, Alice attempts to resolve the issue by following the provided prompts. However, she encounters several complications that hinder her ability to resolve the problem, including an outdated mobile app, and an expired license for the wired connection.
Frustrated with the lack of progress, Alice requests an explanation from Roomie.

Roomie provides a brief explanation, stating that the outdated mobile app and expired license are preventing it from establishing a stable internet connection.
However, this single-shot explanation does not fully satisfy Alice, as she feels she needs a better understanding of how these factors are interconnected and impact Roomie's performance.

To gain a deeper understanding, Alice engages in an argumentation-based dialogue with Roomie.~She presents arguments about the importance of regularly updating the mobile app and renewing the license, citing the need for optimal performance and security.~Roomie counters by explaining that while updates and renewals are important, other factors such as network stability and hardware compatibility also play roles in its ability to function properly.

Through the dialogue, Alice and Roomie explore various aspects of the problem, including the potential risks of using outdated software, the benefits of maintaining a stable power supply, and the importance of regular maintenance. This \textit{dialectical interaction} allows Alice to better understand Roomie's reasoning and the evidence behind its explanations. While she may still have reservations about Roomie's arguments, she now has a more comprehensive grasp of the factors contributing to Roomie's disconnection and can make more informed decisions on how to proceed with troubleshooting.

This example demonstrates how a single-shot reconciliation explanation may not always be sufficient in scenarios requiring deeper understanding. In contrast, an argumentation-based dialogue, such as the one enabled by our proposed framework, allows for a more thorough exploration of the reasoning behind the AI system's behavior, enabling users to gain a more nuanced understanding. We also ran a human user study with this motivating example (see Section~\ref{sec:user-study}) highlighting the strengths of our framework.

\section{Related Work}

The influential work by \citet{walton1995commitment} provides a valuable framework for categorizing dialogues based on participants' knowledge, objectives, and governing rules. This categorization is essential for understanding the distinct characteristics and purposes of different dialogue types. Each dialogue type revolves around a central topic, typically a proposition, that serves as the subject matter of discussion. 

Related dialogue types include:~persuasion \citep{gordon1994,prakken2006formal}, where an agent attempts to convince another agent to accept a proposition they initially do not hold; information-seeking \citep{parsons2003properties,fan2012agent}, where an agent seeks to obtain information from another agent believed to possess it; and inquiry \citep{hitchcock2017some,black2009inquiry}, where two agents collaborate to find a joint proof for a query that neither could prove individually.

While many dialogue systems have been proposed for these dialogue types \cite{black2021argumentation}, to the best of our knowledge, no existing dialogue frameworks have been developed exclusively for model reconciliation processes \cite{chakraborti2017plan}.
This is a crucial aspect of communication that sets our framework apart from related dialogue types, such as persuasion and information-seeking. To better illustrate this, in Section~\ref{semantics}, we provide an example that clarifies the distinctions between our proposed dialogue type and persuasion and information-seeking.

On a similar thread, our work fits well within the literature on argumentation-based explainable AI \cite{ijcai2021p0600}. However, a big difference with most existing approaches within that space \cite{FanT15,collins2019towards,oren2020argument,budan2020proximity,rago2023interactive} is that they are based on forms of \emph{abstract} argumentation, which in our specific setting offers limited expressivity as the internal structure of arguments is ignored. In a practical explanatory dialogue setting with implementations for user studies (such as in our case), one must know and express the contents of the arguments conveyed, and how they can be used to generate new arguments and counterarguments.\footnote{That is why we opted to using deductive argumentation, a form of \emph{structured} argumentation, whose key feature is the clarification of the nature of arguments and counterarguments.}

In similar spirit, \citet{dennis2022explaining} proposed a framework for explaining the behavior of BDI systems. However, the differences lie in the underlying formalisms (BDI vs structure deductive argumentation), and importantly, their methodology lacks an experimental evaluation. In contrast, we include both computational experiments and a human-user study, providing a more robust and empirically grounded understanding of the framework's effectiveness. In an orthogonal direction, \citet{teze2022approach} proposed an argumentation-based approach for epistemic planning that allows for handling contextual preferences of users during plan construction, but without explainability considerations. In contrast, our framework can be used to explain planning problems to users via argumentation-based dialogues.

Finally, our work is motivated by the model reconciliation process (MRP)~\citep{chakraborti2017plan,sreedharan2021using,sreedharan2022explainable}, and specifically the logic-based variant~\citep{son2021model,vas21,vasileioulogic,vasileioua2023please}. Our framework addresses two MRP limitations: (1) the explainer agent's assumed knowledge of the human model (we relax this assumption) and (2) single-shot interactions (we focus on dialogue-based interactions).~Notably, \citet{DungS22a} tackle these limitations using answer set programming, but their approach is tied to planning problems while ours can be used to express general problems. Specifically, our framework relies on the general notion of argument/counterargument, while theirs discuss only arguments related to optimal planning, and it is not clear how to extend it to our general context. Moreover, their framework is purely theoretical and lacks experimental evaluation.

\section{Background: Deductive Argumentation}
\label{sec:background}

We assume familiarity with classical logic and provide a partial review of deductive, logic-based argumentation~\citep{besnard2014constructing}, which serves as the underlying machinery of our proposed framework.

We consider a (propositional) language $\mathcal{L}$ that utilizes the classical entailment relation, represented by $\models$. We use $\bot$ to denote falsity and assume that knowledge bases (finite sets of formulae) are consistent unless specified otherwise.

Our approach relies on an intuitive concept of a logical \emph{argument}, which can be thought of as a set of formulae employed to (classically) prove a particular claim, represented by a formula:

\begin{definition}[Argument]
    Let $\KB$ be a knowledge base and $\phi$ a formula. An \emph{argument} for $\phi$ from $\KB$ is defined as $A = \langle \Gamma, \phi \rangle$ such that: (i) $\Gamma \subseteq \KB$; (ii) $\Gamma \models \phi$; (iii) $\Gamma \not \models \bot$; and (iv) $\nexists \Gamma' \subset \Gamma$ s.t. $\Gamma' \models \phi$.
    \label{def:arg}
\end{definition}

\noindent We refer to $\phi$ as the \emph{claim} of the argument, denoted as $\mathrm{CL}(A)$, and $\Gamma$ as the \emph{premise} of the argument, denoted as $\mathrm{PR}(A)$. The set of all arguments for a claim $\phi$ from $\KB$ is represented by $\mathcal{A}(\KB, \phi)$.

To account for conflicting knowledge between agents, we will make use of a general definition of a \emph{counterargument}, that is, an argument opposing another argument by emphasizing points of conflict on the premises or claim of the argument. With a slight abuse of notation:

\begin{definition}[Counterargument]
   Let $\KB_i$ and $\KB_j$ be two knowledge bases, 
   $A_i = \langle \Gamma_i, \phi_i \rangle$, and $A_j = \langle \Gamma_j, \phi_j \rangle$ be two arguments for $\phi_i$ from $\KB_i$ and for $\phi_j$ from $\KB_j$, respectively. We say that $A_i$ (or $A_j$) is a \emph{counterargument} for $A_j$ (or $A_i$) iff $\Gamma_i \cup \Gamma_j \models \bot$.
\end{definition}

\noindent We denote the set of all counterarguments for an argument $A$ from $\KB$ with $\mathcal{C}(\KB, A)$.

\section{DR-Arg Framework}
\label{drai-frame}

In this section, we introduce the \textit{Dialectical Reconciliation via Structured Argumentative Dialogues} (DR-Arg) framework. We begin by discussing the key assumptions and components of the framework.

\subsubsection{Key Assumptions:}

The DR-Arg framework involves two agents engaging in a dialogue, with one agent taking on the role of an \textit{explainer} (denoted by index~$R$) and the other an \textit{explainee} (denoted by index~$E$). The goal of the dialogue is to \textit{help the explainee understand the decisions made by the explainer from the explainer's perspective}. We use $\phi$ to represent an explainer's decision and $\Phi$ to represent the set of all decisions the explainee seeks to understand. 

Three critical assumptions underlie our framework:

\begin{enumerate}

    \item \textbf{Agent Knowledge Bases:} The explainer is associated with a knowledge base $\KB_R$ that encodes its own knowledge of the underlying task. The explainee is associated with knowledge base $\KB_E$ that encodes \emph{their approximation of the explainer's knowledge}, which can be $\emptyset$. No agent has explicit access to the other agent's knowledge base.

    \item \textbf{Explainee Queries:} Initiated by the explainee, the dialogue starts with a \texttt{query} $\phi \in \Phi$, where $\KB_E \not \models \phi$ (or $\KB_E \models \neg \phi$) and $\KB_R \models \phi$. The explainee has the flexibility to generate subsequent queries dynamically as the dialogue progresses, reflecting their evolving understanding and the need for additional clarification.

    \item \textbf{Public Commitment Stores:} Both agents contribute to public \textit{commitment stores} that store their utterances throughout the dialogue, akin to a ``chat log". A commitment store for agent $x\in \{R, E\}$ is defined as $CS_x = ( CS_x^1, \ldots, CS_x^t )$, where $CS_x^t = \langle l(\gamma), A \rangle$ and $l(\gamma)$ is an instantiated locution (see next section) and $A$ the respective argument (can be empty) accompanying the locution. This feature allows to build more complex and contextually aware arguments.
\end{enumerate}

\noindent The main goal of the DR-Arg is formulated as follows:

\begin{quote}
    \emph{Given an explainer agent with $\KB_R$, an explainee agent with $\KB_E$, and a set of queries $\Phi$ such that, for all $\phi \in \Phi$, $\KB_E \not \models \phi$ (or $\KB_E \models \neg \phi$) and $\KB_R \models \phi$, the goal of DR-Arg is to \emph{enable} $\KB_E \models \phi$ through dialectical reconciliation.}
\end{quote}

A critical aspect of this formulation is successfully \textit{enabling} $\KB_E \models \phi$ during the dialogue between explainee and explainer. At a high level, we aim to find a way to help the explainee transition from a state of \textit{not understanding} a decision $\phi$ (i.e.,~$\KB_E \not \models \phi$ or $\KB_E \models \neg \phi$) to a state of \textit{understanding} the decision (i.e.,~$\KB_E \models \phi$). Our thesis is that a natural way of achieving this transition is through an argumentation-based dialogue that facilitates dialectical reconciliation, i.e., a \textit{dialectical reconciliation dialogue}.

At a high level, a dialectical reconciliation dialogue is a process resolving inconsistencies, misunderstandings, and knowledge gaps between the explainer and the explainee. This is achieved through argument exchange and dialogue moves that collaboratively construct a shared understanding of the explainer's decisions. To successfully achieve a dialectical reconciliation dialogue, the agents should follow certain (dialogue) protocols that guide their interaction:

\squishlist
\item Establish a clear dialogue structure, including the use of \emph{locutions} that define permissible speech acts and turn-taking mechanisms.
\item Engage in a cooperative and collaborative manner, with both agents focusing on the shared goal of improving the explainee's understanding.
\item Employing argumentation techniques, such as offering counterexamples or pointing out logical inconsistencies, to constructively challenge each other's positions.

\squishend

Following these protocols, the explainer helps the explainee iteratively refine their knowledge base, ultimately converging on a shared understanding that enables $\KB_E \models \phi$ for all decisions $\phi \in \Phi$.

\begin{table*}[!t]
 \centering  
 \renewcommand{\arraystretch}{2.1}
\resizebox{1.65\columnwidth}{!}{ 
\begin{tabular}{ |c|c|c|c| } 
\hline
\textbf{Locution} & \textbf{Agent Type} &\textbf{Preconditions} & \textbf{Effects} \\
\hline
\hline

$\texttt{query}(\gamma)$  & $E$ & \makecell[l]{(1) $\exists A \in CS_R^T$ s.t. $\gamma \subseteq \mathrm{PR}(A)$ \textbf{and} \\
(2) $\texttt{query}(\gamma)\not \in CS_E^T$ \textbf{and}  \\ 
(3) $\KB_E \not \models \gamma$ or $\KB_E \models \neg \gamma$} 
& $CS^t_E \gets  \langle \texttt{query}(\gamma), \emptyset \rangle$ \\
\hline

$\texttt{support}(\gamma)$  & $R$ & \makecell[l]{(1) $\texttt{query}(\gamma)\in CS_E^{t-1}$ \textbf{and} \\
(2) $\exists A \in \mathcal{A}(\KB_R, \gamma)$ s.t. $A \not \in CS_R^T$} 
& $CS^t_R \gets \langle \texttt{support}(\gamma), A \rangle$ \\
\hline

\multirow{2}{*}{$\texttt{refute}(\gamma)$}  & $E$ & \makecell[l]{(1) $\exists A \in CS_R^{T}$ s.t. $\gamma \subseteq \mathrm{PR}(A) \cup \mathrm{CL}(A)$ \textbf{and} \\
(2) $\exists A \in \mathcal{C}(\KB_E \cup CS_R^{T}, \gamma)$ s.t. $A\not \in CS_E^T$} 
& $CS^t_E \gets \langle \texttt{refute}(\gamma), A \rangle$ \\
\cline{2-4}
  & $R$ & \makecell[l]{(1) $\exists A \in CS_E^{T}$ s.t. $\gamma \subseteq \mathrm{PR}(A) \cup \mathrm{CL}(A)$ \textbf{and} \\
(2) $\exists A \in \mathcal{C}(\KB_R \cup CS_E^{T}, \gamma)$ s.t. $A\not \in CS_R^T$} 
& $CS^t_R \gets \langle \texttt{refute}(\gamma), A \rangle$ \\
\hline

\multirow{2}{*}{\texttt{understand}} & $E$ & \makecell[l]{(1)  $\texttt{query}(\gamma)$ preconditions do not hold  \textbf{and} \\
(2) $\texttt{refute}(\gamma)$ preconditions do not hold}
&$CS^t_E \gets \langle \texttt{understand}, \emptyset \rangle$  \\
\cline{2-4}
 & $R$ & \makecell[l]{(1)  $\texttt{support}(\gamma)$ preconditions do not hold  \textbf{and} \\
(2) $\texttt{refute}(\gamma)$ preconditions do not hold }
& $CS^t_R \gets \langle \texttt{understand}, \emptyset \rangle$  \\

\hline
\end{tabular}
}
\caption{The DR dialogue protocol. Note that, with a slight abuse of notation, the condition $A \in CS_x^T$ ($x \in \{R, E\}$) is true if there exists an argument $A$ that has been uttered by agent $x$ at any step during the dialogue, i.e.,  $1 \leq T \leq t-1$.} 
\label{DR_semantics}
\end{table*}

\subsection{Dialectical Reconciliation Dialogue Type}

We now formalize the dialectical reconciliation dialogue type, inspired by Hamblin's dialectical games framework \citep{hamblin1970fallacies,hamblin1971mathematical}. Here, a dialogue is viewed as a game-theoretic interaction, where utterances are treated as moves governed by rules that define their applicability. In this context, moves consist of a set of \emph{locutions}, which determine the types of permissible utterances agents can make. To align with the goal of DR-Arg, we define the following set of locutions:
\begin{equation}
L = \{\texttt{query}, \texttt{support}, \texttt{refute}, \texttt{understand} \}
\end{equation}

The \texttt{query} locution enables the explainee to ask the explainer for an argument supporting the explainee's \texttt{query}. The $\texttt{support}$ locution allows the explainer to provide a supporting argument for the explainee's \texttt{query}. The $\texttt{refute}$ locution permits both agents to provide counterarguments, and the \texttt{understand} locution allows both agents to acknowledge each other's utterances when no further queries or counterarguments are possible. We impose two restrictions: (1)~the $\texttt{query}$ locution is only available to the explainee, and (2)~the $\texttt{support}$ locution is only available to the explainer. These restrictions are reasonable given the goal of DR-Arg; future work will explore relaxing them.

Note that we opted for an \texttt{understand} locution instead of a simple \texttt{agree} (or \texttt{accept}) locution as the goal of DR-Arg is not to convince the explainee about $\Phi$ but to help them understand $\Phi$.~An \texttt{understand} locution reflects this flexibility, where agents do not have to agree with each other; they only have to acknowledge each other's utterances and understand each other's perspectives. 

Locutions are typically instantiated with specific formulae that make up the range of possible \emph{dialogue moves} $m_t$:
\begin{equation}
m_t = \langle x, l(\gamma) \rangle,
\end{equation}

\noindent where $t$ is an index indicating the dialogue timestep, $x\in \{R, E\}$ denotes the agent making the move, $l\in L$ is a locution, and $\gamma \in \mathcal{L}$ is a formula that instantiates the locution (e.g.,~the content of the move). 

We now formally define a dialectical reconciliation (DR) dialogue. A DR dialogue requires that the first move must always be a $\texttt{query}$ locution from the explainee, and the agents take turns making and receiving moves:

\begin{definition}[DR Dialogue]
A \emph{DR dialogue} $D$ is a sequence of moves $[m_1,\ldots, m_{|D|}]$ involving an explainee agent $E$ and an explainer agent $R$, where the following conditions hold:
\squishenum
\item[1.] $m_1 = \langle E, \texttt{query}(\phi) \rangle$ is the opening move of the dialogue made by the explainee.
\item[2.] Each agent can make only one move $m_t$ per (alternating) timestep $t$.
\squishend
\label{DR_dialogue_def}
\end{definition}

\noindent We refer to the initial \texttt{query} $\phi$ as the \textit{starting topic} of the dialogue, and to all explainee queries $\Phi$ made in the dialogue as the \emph{overall topic} of the dialogue.

A DR dialogue is \emph{terminated} at timestep $t$ if and only if the explainee cannot generate subsequent queries or counterarguments, that is, when the explainee utters the \texttt{understand} locution. More formally,

\begin{definition}[Terminated DR Dialogue]
\label{def:term}
    A DR dialogue $D$ is \emph{terminated} at timestep $t$ iff $m_t = \langle E, \texttt{understand} \rangle$ and $\nexists t'<t$ s.t. $D$ is terminated at timestep $t'$.
\end{definition}

\subsubsection{Agent Strategy:} 
During the dialogue, the agents essentially determine their moves based on objectives like adhering to rationality or influencing dialogue length. In other words, each agent follows a \emph{strategy} when selecting their next move. For an agent $x$, a strategy, denoted $S_x$, is a function taking in its current dialogue $D$, knowledge base $\KB_x$, and next timestep $t$ to output the next move. 

While strategies can take several forms (e.g.,~preference-based, probabilistic), for simplicity, we assume two ordered strategies:~$S_E(D, \KB_E, t) = [\texttt{refute}$, $\texttt{query}$, $\texttt{understand}]$~and~$S_R(D, \KB_R, t) = [\texttt{support}$, $\texttt{refute}$, $\texttt{understand}]$, where the ordered lists show the priorities of dialogue moves for the explainee and explainer, respectively, at $t > 1$.

Now, if the agents follow their respective strategies during the DR dialogue, and the dialogue does not continue after it has terminated, then we say that the dialogue is \emph{well-formed}.

\begin{definition}[Well-Formed DR Dialogue]
A DR dialogue $D$ is \emph{well-formed} iff it is terminated at timestep $t$ and, for all timesteps $1<t'<t$, $m_{t'}\in S_x(D', \KB_x, t')$ for each move $m_{t'}$ from agent $x$, where $D' \subseteq D$ consists of the first $|D'| = t' - 1$ moves from $D$.
\end{definition}

\subsection{Operational Semantics of DR Dialogues}
\label{semantics}

In argumentation-based dialogues, the combination of locutions and formulae by agents is not arbitrary; rather, it is governed by specific rules. This restriction is encapsulated in the concept of a \emph{dialogue protocol}. A dialogue protocol delineates the \emph{operational semantics} of a dialogue, explicating the preconditions and effects for each locution \citep{plotkin1981structural}. That is, locutions exhibit action-like properties, influencing and modifying the state of the dialogue.

As described in Definition~\ref{DR_dialogue_def}, the dialogue is initiated with a $\texttt{query}$ move from the explainee ($m_1$). Recall also that the $\texttt{query}$ and $\texttt{support}$ locutions are restricted to the explainee and explainer, respectively. Table~\ref{DR_semantics} describes the generation of valid dialogue moves $m_t$ ($t>1$) during a DR dialogue. 

A $\texttt{query}$ locution with formula $\gamma$ is valid if it satisfies three preconditions: (1) $\gamma$ is part of the premise in an argument previously made by the explainer, (2) $\gamma$ has not been queried before, and (3) $\gamma$ is neither entailed by $\KB_E$ nor is its negation entailed. The $\texttt{support}$ locution, instantiated with formula $\gamma$, is permissible when: (1) $\gamma$ was queried by the explainee in the preceding timestep, and (2) a new argument for $\gamma$ exists in $\KB_R$. The $\texttt{refute}$ locution is instantiated with $\gamma$ if: (1) $\gamma$ is in the premises or claim of any argument made by either the explainer (resp. explainee), and (2) an unasserted counterargument refuting $\gamma$ exists in $\KB_E$ (resp. $\KB_R$). The $\texttt{understand}$ locution is a valid option if $\texttt{query}$ (resp. $\texttt{support}$) and $\texttt{refute}$ cannot be uttered by the explainee (resp. explainer). After each move, the respective agents' commitment stores are updated.

Note that our framework remains neutral regarding to which argument ($\texttt{support}$ move) or counterargument ($\texttt{refute}$ move) is computed first. This can be done in a preference-based fashion by incorporating and minimizing a cost function that measures the complexity of the arguments. For simplicity again, we employ a cost function based on argument length, i.e., $cost(A) = |\mathrm{PR}(A)|$.

Importantly, our framework permits agents to utilize each other's commitment stores when formulating arguments, specifically for the \texttt{refute} locution (see precondition (2)). This inter-use of commitment stores enables the agents to draw upon shared information to construct arguments, thereby creating a more realistic representation of dialectical reconciliation.

\subsection{Illustrative Example}
\label{example}

Consider the following explainer and explainee knowledge bases, where all formulae are equally preferred:
\begin{align*}
      \KB_R &= \{ a, b , a \land b \!\rightarrow\! c, d, d\!\rightarrow\!\neg e, f, f \!\rightarrow\! d\} \\
    \KB_E &= \{e, e \!\rightarrow\! \neg c, g, g \land a\!\rightarrow\! \neg f \}
\end{align*}

\noindent The starting topic is $c$, where $\KB_R \models c$ and $\KB_E \models \neg c$.

A generated DR dialogue is shown in Table~\ref{DR_example}. The dialogue begins with the explainee asking the explainer about $c$ ($m_1$), and the explainer provides an argument \texttt{support}ing it ($m_2$). The explainee counters by refuting $c$ with $e$ ($m_3$), which the explainer then \texttt{refute}s with $d$ ($m_4$). Next, the explainee poses a new \texttt{query} about $d$ ($m_5$), and the explainer \texttt{support}s it with $f$ ($m_6$). The explainee subsequently \texttt{refute}s $f$ with $g$ and $a$ (from the explainer's commitment store) ($m_7$). Finally, both agents utter \texttt{understand} ($m_8$ and $m_9$), leading to the termination of the dialogue.

\begin{table}
\renewcommand{\arraystretch}{1.2}
\resizebox{1.0\columnwidth}{!}{ 
\begin{tabular}{ l l } 
 \textbf{Dialogue Move} & \textbf{Commitment Store} \\
\hline

 $m_1 = \langle E, \texttt{query}(\{c\}) \rangle$ & $CS_E^1 = \langle \texttt{query}(\{c\}), \emptyset \rangle$ \\ 
 $m_2 = \langle R, \texttt{support}, \{c\} \rangle$ & $CS_R^2 = \langle \texttt{support}(c), \langle \{ a, b, a \land b\!\rightarrow\! c \}, c \rangle \rangle$ \\ 
 $m_3 = \langle E, \texttt{refute}(\{c\}) \rangle$ & $CS_E^3 = \langle \texttt{refute}(\{c\}),
    \langle \{ e, e \!\rightarrow\! \neg c \}, \neg c \rangle \rangle$ \\
 $m_4 = \langle R, \texttt{refute}(\{e\}) \rangle$ & $CS_R^4 = \langle \texttt{refute}(\{e\}), \langle \{ d, d\!\rightarrow\! \neg e\}, \neg e \rangle \rangle$ \\
 $m_5 = \langle E, \texttt{query}(\{d\}) \rangle$ & $CS_E^5 = \langle \texttt{query}(\{d\}), \emptyset \rangle$ \\
 $ m_6 = \langle R, \texttt{support}(\{d\}) \rangle$ & $ CS_R^6 = \langle \texttt{support}(\{d\}), \langle \{ f, f \!\rightarrow\! d\}, d \rangle \rangle $ \\
 $m_7 = \langle E, \texttt{refute}(\{f\}) \rangle$ & $CS_E^7 = \langle \texttt{refute}(\{f\}), \langle \{ g, a, g \land a \!\rightarrow\! \neg f\}, \neg f \rangle \rangle$ \\
 $ m_8 = \langle R, \texttt{understand} \rangle$ & $CS_R^8 = \langle \texttt{understand}, \emptyset \rangle$ \\
 $m_9 = \langle E, \texttt{understand} \rangle$ & $CS_E^9 = \langle \texttt{understand}, \emptyset \rangle$ \\

\end{tabular}
}
\caption{Example of DR dialogue.}
\label{DR_example}
\end{table}

It is important to note that the goal we pursue in this work (dialectical reconciliation) sets our framework apart from traditional argumentation frameworks that aim to achieve mutual agreement through persuasion \cite{gordon1994,prakken2006formal,parsons2003properties} or obtain information through information-seeking \cite{parsons2003properties,fan2012agent}. To better highlight the differences, let us consider the 
logic-based persuasion and information-seeking frameworks presented in \cite{parsons2002analysis,parsons2003properties}. In these frameworks, agents are assumed to have ``dialogical attitudes'' (akin to agent strategies) when choosing their assert and accept moves. The attitudes relevant to our setting are the confident agent, who asserts any proposition for which an argument can be constructed, and the cautious agent, who accepts a proposition only if they cannot construct a counterargument against it.
In our example, 
given that the starting dialogue topic is $c$, the goal in persuasion is for the explainer to persuade the explainee to accept $c$, while in information-seeking, the explainee aims to gather information about $c$. The 
corresponding dialogues are shown in Table~\ref{P_IS_example}.

\begin{table}[h]
\renewcommand{\arraystretch}{1.2}
\resizebox{1.0\columnwidth}{!}{ 
\begin{tabular}{ l  l } 
 \textbf{Persuasion} & \textbf{Information-seeking} \\
\hline

$m_1  = \langle R, \texttt{assert}(c) \rangle$ & $m_1  = \langle E, \texttt{question}(c) \rangle$\\
$m_2  = \langle E, \texttt{assert}(\lnot c) \rangle$ & $m_2  = \langle R, \texttt{assert}(c) \rangle$ \\
$m_3  = \langle R, \texttt{challenge}(\lnot c) \rangle$ & $m_3  = \langle E, \texttt{challenge}(c) \rangle$\\
$m_4  = \langle E, \texttt{assert}(\{e, e\rightarrow \lnot c\}) \rangle$ & $m_4  = \langle R, \texttt{assert}(\{a, b, a \land b\rightarrow c\}) \rangle$ \\
$m_5  = \langle R, \texttt{assert}(\lnot e) \rangle$ & $m_5  = \langle E, \texttt{accept}(\{a\} \rangle$ \\

$m_6  = \langle E, \texttt{assert}(e) \rangle$ & $m_6  = \langle E, \texttt{accept}(\{b\} \rangle$\\

$m_7  = \langle R, \texttt{challenge}(e) \rangle$ & $m_7  = \langle E, \texttt{accept}(\{a \land b \rightarrow c\} \rangle$ \\
$m_8  = \langle E, \texttt{assert}(\{e\}) \rangle$ & \\

\end{tabular}
}
\caption{Example of persuasion and information-seeking dialogues.}
\label{P_IS_example}
\end{table}

The differences between a DR dialogue and persuasion and information-seeking dialogues are 
evident in this example. Compared to persuasion, the primary difference lies in the goal. A DR dialogue aims for understanding, while persuasion seeks to change the explainee's beliefs. This is evident in the dialogue moves, where dialectical reconciliation allows for a back-and-forth exchange of arguments and counterarguments ($m_3$ to $m_8$) until a point of understanding is reached ($m_9$). In persuasion, the dialogue ends when the explainer concedes ($m_8$), failing to persuade $E$ about $c$.

Compared to information-seeking, the main difference is the level of interaction. A DR dialogue enables the explainee to provide counterarguments (e.g., refuting $c$ in $m_3$) and the explainer to offer additional information (e.g., $m_4$ onwards). This rich exchange is not possible in the information-seeking protocol, where the explainee simply accepts the explainer's assertions ($m_4$) without the opportunity to challenge or seek further clarification.

This simple example shows that a DR dialogue provides a more interactive and collaborative framework for understanding, compared to the one-sided nature of persuasion and the limited interaction in information-seeking.

\subsection{Properties of DR Dialogues}
\label{properties}

We now describe two properties for assessing the efficacy of a DR dialogue: \emph{termination} and \emph{success}. 

\subsubsection{Termination:} This property ensures that the dialogue concludes within a finite number of steps and is devoid of any deadlocks, guaranteeing that at every stage, each agent has at least one viable move.

\begin{theorem}
    Every DR dialogue is guaranteed to terminate.
    \label{term_theo}
\end{theorem}

\begin{sproof}
First, the operational semantics (see Table~\ref{DR_semantics}) outline the constraints and conditions under which each dialogue move can be executed. Second, the agents' knowledge bases are finite, meaning that there are only a limited number of different moves that can be generated, and the agents cannot repeat these moves. As such, the dialogue will not continue indefinitely. 

We now prove through contradiction that a deadlock cannot happen. Assume that a deadlock happened, where an agent $x$ does not have any available moves to make and the dialogue has not terminated. There are two cases:
\squishlist
\item Agent $x$ is an explainee. When the explainee cannot make any $\texttt{query}$ or $\texttt{refute}$ moves, it can always make the \texttt{understand} move since its preconditions are that the preconditions of the $\texttt{query}$ and $\texttt{refute}$ moves do not hold. 
\item Agent $x$ is an explainer.~Similar to the previous case, when the explainer cannot make any $\texttt{support}$ or $\texttt{refute}$ moves, it can always make the \texttt{understand} move.
\squishend
This contradicts our assumption and the dialogue is thus deadlock-free. Therefore, a DR dialogue is guaranteed to terminate.
\end{sproof}

\subsubsection{Success:} The success of a terminated dialogue is contingent upon the achievement of its primary goal. For DR-Arg, this entails the explainee comprehending the overall topic $\Phi$, from the explainer agent's perspective. This is formally denoted as $\KB_E \models \phi$ for each $\phi \in \Phi$, or more succinctly, $\KB_E \models \Phi$. Achieving this involves a \emph{knowledge update} in $\KB_E$, incorporating the explainer's arguments from the dialogue. We adopt the following general knowledge base update from the literature \citep{vasileioulogic}:

\begin{definition}[Updated Knowledge Base]
    The \emph{updated knowledge base} $\KB_E$ upon integrating argument $A$ is defined as $\widehat{\KB}_E^{A} = (\KB_E \cup \mathrm{PR}(A))\setminus M$, where $M \subseteq \KB_E \setminus \mathrm{PR}(A)$ is a $\subseteq$-minimal subset whose (potential) removal ensures that $(\KB_E \cup \mathrm{PR}(A))$ remains consistent.
    \label{def:kb_update}
\end{definition}

For simplicity, we assume that the knowledge base update transpires post-dialogue. Performing this update during the dialogue is equally feasible, given that the explainee has access to the explainer's commitment store, which aids in formulating new arguments. This means that the timing of the update does not affect the argumentation dynamics.

Now, a crucial observation is that not all arguments presented by the explainer are necessary to update $\KB_E$ for it to entail $\Phi$. An incremental update strategy can be employed, beginning with the most recent argument and proceeding until $\KB_E \models \Phi$ is fulfilled. Should retraction be needed for consistency, it is confined to the original contents of $\KB_E$, preserving the integrity of the added arguments. This approach assures that $\KB_E \models \Phi$ is enabled. Hence, a DR dialogue that attains its objective is deemed \emph{successful}.

\begin{definition}[Successful DR Dialogue]
    A terminated DR dialogue $D$ regarding topic $\Phi$ is \emph{successful} iff $\widehat{\KB}_E^{A} \models \Phi$ for some $A \subseteq CS_R$.
    \label{def:succ}
\end{definition}

Integrating Definition~\ref{def:succ} with the underlying principles of the DR-Arg framework leads to an important conclusion:

\begin{theorem}
    A terminated DR dialogue $D$ on topic $\Phi$ is always successful.
    \label{succ_theo}
\end{theorem}

\begin{sproof}
First, recall that the topic of the dialogue $\varphi$ must be entailed by the explainer (i.e.,~$KB_R \models \varphi$), which means that an argument for $\varphi$ from $KB_R$ always exists (Definition~\ref{def:arg}).

Now, notice that for a terminated dialogue $D$, the explainer's commitment store $CS_R$ contains the explainer's set of arguments that have been presented during the dialogue. Since $KB_R \models \varphi$, and the arguments in $CS_R$ are derived from $KB_R$, it follows that using the arguments in $CS_R$ to update the explainee's knowledge base $KB_E$ (w.r.t. Definition~\ref{def:kb_update}) will enable $KB_E \models \varphi$, as in the worst case, the entire $CS_R$ will be used to update $KB_E$.

Therefore, the explainee's knowledge base will eventually entail $\varphi$ (i.e.,~$KB_E \models \varphi$) and, as such, a terminated DR dialogue on topic $\varphi$ is always successful.   
\end{sproof}

\section{Approximating Explainee Understanding}
\label{understand}

\textit{Understanding}, a multifaceted and abstract concept, is challenging to quantify and often involves the explainee’s cognitive process of forming a functional mental model of the subject matter, which includes its causes, consequences, and interconnections. This process resembles constructing a complex ``blueprint'' through the narrative provided by the explainer, effectively facilitated by argumentation-based dialogue. Such dialogues engage the explainee’s cognitive abilities, enhancing learning and understanding through active engagement, reconstruction, and assimilation of information, as evidenced in cognitive psychology studies~\citep{johnson1983mental,mercier2011humans}. Our framework is motivated by these insights, employing argumentation to guide the explainee in developing a comprehensive understanding of the phenomenon under discussion.

As stated, our main objective is to enhance the explainee's understanding of the explainer's decisions. To quantify and approximate this understanding, we propose a simple metric that measures the similarity between the explainee's knowledge base ($\KB_E$) and the explainer's knowledge base ($\KB_R$). We postulate that \textit{the explainee's understanding is likely to improve as the similarity between $\KB_E$ and $\KB_R$ increases}.

We define the similarity between $\KB_E$ and $\KB_R$ using syntactic and semantic measures. Syntactic similarity assesses structural likeness (e.g., similarity of formulae), while semantic similarity examines the logical consequences of the knowledge bases. We employ a weighted Sørensen-Dice similarity index~\citep{dice1945measures,sorensen1948method} as follows:

\begin{equation}
    \Sigma = a \cdot \frac{2\cdot|\KB_E \cap \KB_R|}{|\KB_E
|+|\KB_R|} +(1-a)\cdot \frac{2\cdot|\fml{B}_E \cap \fml{B}_R|}{|\fml{B}_E|+|\fml{B}_R|}
\end{equation}
\noindent where $a \in [0,1]$ is a parameter indicating the weight of each metric component. Here, $\fml{B}_E$ and $\fml{B}_R$ represent the backbone literals of $\KB_E$ and $\KB_R$, respectively, which are the literals entailed by each knowledge base \citep{parkes1997clustering}.\footnote{Note that instead of the backbone literals of the knowledge bases, we could alternatively consider their prime implicates, which are their strongest consequences \cite{jackson1992computing}.} This formula approximates the explainee's level of understanding as the similarity between $\KB_E$ and $\KB_R$.

Note that we assume that the explainee's knowledge base is dynamic, capable of assimilating new information from the explainer. We also assume that the explainee, as a rational agent, actively seeks to understand the explainer's perspective and integrates this information into $\KB_E$ (Definition~\ref{def:kb_update}).\footnote{Recall that $KB_E$ is what the explainee thinks the agent's knowledge is, which means that they have no qualms adopting information from $KB_R$.}

\begin{example}
Consider the DR dialogue from the illustrative example. Upon dialogue termination, the explainee sequentially updates $\KB_E$ with the explainer's arguments until the dialogue topic $\Phi = \{c, d\}$ is entailed by $\KB_E$ (i.e., $\KB_E \models c$ and $\KB_E \models d$). Table~\ref{example-KB_update} illustrates the evolution of the knowledge base similarity with each update.

\begin{table}[!h]
\renewcommand{\arraystretch}{1.8}
\setlength{\tabcolsep}{3pt}
\resizebox{1.0\columnwidth}{!}{ 
\begin{tabular}{c l c l } 

\textbf{\#} & \textbf{Premise to Add} & \multirow{1}{*}{\textbf{Updated $\KB_E$}} & \textbf{Similarity Metric} \\
\hline

1 & $\{f, f\!\rightarrow\! d \}$ & $\{e, e\!\rightarrow\! \neg c, g,f, f \!\rightarrow\! d \}$& $\Sigma = 0.5 \!\cdot\! \frac{2\cdot 2}{12}  \!+\! 0.5 \!\cdot\! \frac{2\cdot 2}{11} = 0.35$ \\

2 & $\{d, d\!\rightarrow\! \neg e \}$ & $\{e\!\rightarrow\! \neg c, g, f, f \!\rightarrow\! d, d, d\!\rightarrow\! \neg e\}$  & 
$\Sigma = 0.5 \!\cdot\! \frac{2\cdot 4}{13}  \!+\!  0.5 \!\cdot\! \frac{2\cdot 3}{11} = 0.58$ \\

3 & $\{a, b, a\land b\!\rightarrow\! c\}$ & \makecell{$\{ e\!\rightarrow\! \neg c, g, f, f \!\rightarrow\! d,d, d\!\rightarrow\! \neg e,$ \\ $ a, b, a\land b \!\rightarrow\! c \}$} & $\Sigma = 0.5 \!\cdot\! \frac{2\cdot 7}{16}  \!+\!  0.5 \!\cdot\! \frac{2 \cdot 6}{13} = 0.90$ 
\end{tabular}
}
\caption{Example of knowledge base update and similarity metric.}
\label{example-KB_update}
\end{table}

\noindent It is interesting to see how this example underscores the potential advantage of dialectical reconciliation over a single-shot reconciliation approach. For instance, using the single-shot reconciliation approach by \citet{vasileioulogic}, we get the explanation tuple $\fml{E} = \langle \fml{E}^+, \fml{E}^- \rangle = \langle \{ a, b, a\land b \!\rightarrow\! c \}, \{e\} \rangle$, where $\fml{E}^+$ and $\fml{E}^-$ denote the formulae to be added and retracted from $\KB_E$, respectively. Updating $\KB_E$ with $\fml{E}$ (using Definition~\ref{def:kb_update}) results in $KB_E = (\KB_E \cup \fml{E}^+)\setminus \fml{E}^- = \{e \rightarrow \lnot c, g, g\land a \rightarrow \lnot f, a, b, a \land b \rightarrow c \}$. Calculating the similarity score between this updated $\KB_E$ and $\KB_R$, we get $\Sigma = 0.50$. Unsurprisingly, the single-shot reconciliation approach yields a lower similarity score than dialectical reconciliation.
\end{example}

\begin{table*}
\renewcommand{\arraystretch}{1.2}
  \setlength{\tabcolsep}{4pt}
   \centering
 \resizebox{1.99\columnwidth}{!} {
  \begin{tabular}{|c||rrrrr|rrrrr|rrrrr|rrrrr|}
  \hline
   \multirow{2}{*}{$|KB|$} & \multicolumn{5}{c|}{$c = 0.2$} & \multicolumn{5}{c|}{$c=0.4$} & \multicolumn{5}{c|}{$c=0.6$} & \multicolumn{5}{c|}{$c=0.8$} \\
    \multicolumn{1}{|c||}{} & \multicolumn{1}{c}{$T$} & $L$ & $N$ & \multicolumn{1}{c}{$\Delta \Sigma_{\text{\emph{DR}}}$} & \multicolumn{1}{c|}{$\Delta \Sigma_{\text{\emph{SSR}}}$} & \multicolumn{1}{c}{$T$} & $L$ & $N$ & \multicolumn{1}{c}{$\Delta \Sigma_{\text{\emph{DR}}}$} & \multicolumn{1}{c|}{$\Delta \Sigma_{\text{\emph{SSR}}}$} & \multicolumn{1}{c}{$T$} & $L$ & $N$ & \multicolumn{1}{c}{$\Delta \Sigma_{\text{\emph{DR}}}$} & \multicolumn{1}{c|}{$\Delta \Sigma_{\text{\emph{SSR}}}$} & \multicolumn{1}{c}{$T$} & $L$ & $N$ & \multicolumn{1}{c}{$\Delta \Sigma_{\text{\emph{DR}}}$} & \multicolumn{1}{c|}{$\Delta \Sigma_{\text{\emph{SSR}}}$}  \\
  \hline
  \hline
  $2 \times 10^2$ & 0.05s & 21 & 5 & $11.50\%$ &$9.00\%$ & 0.04s & 11 & 1 &  $10.10\%$ &$9.20\%$ & 0.02s & 9 & 2 & $9.90\%$ &$9.20\%$ & 0.05s & 9 & 2 & $9.95\%$ &$9.10\%$  \\
  
  $4 \times 10^2$ & 0.07s & 15 & 6 & $4.50\%$ &$2.50\%$ &0.07s & 15 & 6 & $5.20\%$ &$4.76\%$ & 0.05s & 11 & 5 & $5.63\%$ &$4.19\%$ & 0.06s & 11 & 5 & $5.60\%$ &$5.30\%$ \\
  
  $6 \times 10^2$ & 0.10s & 11 & 5 & $2.83\%$ &$1.37\%$ & 0.10s & 11& 5& $2.15\%$ &$1.43\%$ & 0.20s & 23  & 11 & $4.27\%$ &$1.58\%$ & 0.40s & 59 & 29 & $11.57\%$ &$1.92\%$  \\
  
  $8 \times 10^2$ & 0.30s & 41& 16 & $5.09\%$ &$0.80\%$ & 0.40s & 43 & 20 & $6.45\%$ &$0.74\%$ & 0.40s & 43 & 9 & $3.47\%$ &$0.73\%$ & 0.50s & 43 & 8 & $3.50\%$ &$0.72\%$ \\
\hline 
\hline
$2 \times 10^3$ &  0.50s & 5 & 2 & $0.53\%$ &$0.83\%$ & 1.00s & 23 & 9 & $2.50\%$ &$0.50\%$ & 2.40s&  69& 31 & $5.48\%$ &$0.45\%$ & 1.10s & 25 & 10 & $3.57\%$ &$0.72\%$  \\
  
  $4 \times 10^3$ & 4.30s & 61 & 29 & $4.88\%$ &$0.37\%$ & 5.50s & 71 & 34 & $6.05\%$ &$1.43\%$ & 10.20s & 109 & 54 & $6.72\%$ &$0.59\%$ & 8.50s & 85 & 42 &$6.37\%$ &$1.73\%$     \\
  
  $6 \times 10^3$ & 3.50s & 13 & 6 & $0.89\%$ &$0.20\%$ & 113.00s & 87 & 40 & $4.65\%$ &$0.18\%$ & 3.70s & 13 & 6 & $3.03\%$ &$0.24\%$ & 8.30s & 57 & 28 & $4.93\%$ &$0.23\%$ \\
  
 $8 \times 10^3$ & 7.60s & 43 & 21 & $3.30\%$ &$1.53\%$ & 5.70s & 19 & 9 & $4.03\%$ &$2.86\%$ & 37.90s & 43 & 21 & $5.13\%$ &$4.18\%$ & 5.60s & 19 & 9 & $4.45\%$ &$4.19\%$  \\

\hline 
\hline
  $2 \times 10^4$ & 21.20s & 9 & 4 & $0.88\%$ &$0.15\%$ & 21.70s & 9 & 4 & $0.10\%$ &$0.75\%$ & 21.60s& 9 & 4 & $2.25\%$ &$0.68\%$ & 21.70s& 9 & 4& $2.49\%$ &$0.07\%$  \\
  
  $4 \times 10^4$ & 38.40s& 44& 17 & $3.20\%$ &$1.95\%$ & 45.50s& 66 & 18& $4.30\%$ &$2.13\%$ & 50.20s& 61 & 16 & $5.40\%$ &$4.19\%$ & 55.80s& 68& 23& $6.20\%$ &$3.32\%$    \\
  
  $6 \times 10^4$ & 125.30s & 90 & 33 & $9.40\%$ &$7.31\%$ & 133.00s& 111& 52& $29.40\%$ &$5.15\%$ & 129.60s& 101& 48 & $33.20\%$ &$17.20\%$ &141.50s & 120& 61& $44.90\%$ &$21.32\%$ \\
  
  $8 \times 10^4$ & 149.00s & 95 & 32& $15.60\%$ &$4.79\%$ &155.00s & 129 & 59& $25.40\%$ &$13.41\%$ &161.50s & 121& 42& $30.10\%$ &$21.29\%$ & 172.50s& 155& 72 & $39.30\%$ & $19.47\%$ \\
\hline
\hline
$2 \times 10^5$ & 220.20s & 159 & 63 & $20.30\%$ & $13.14\%$ & 232.50s & 191 & 82 & $32.00\%$ & $22.08\%$ & 242.00s & 202 & 95 & $39.00\%$ & $25.50\%$ & 254.80s & 233 & 108 & $50.20\%$ & $25.59\%$ \\
$4 \times 10^5$ & 386.60s & 245 & 111 & $28.10\%$ & $15.48\%$ & 411.30s & 287 & 135 & $37.90\%$ & $29.76\%$ & 430.00s & 306 & 151 & $45.10\%$ & $32.70\%$ & 456.60s & 340 & 168 & $57.40\%$ & $31.00\%$ \\
$6 \times 10^5$ & 561.20s & 322 & 151 & $33.80\%$ & $19.29\%$ & 594.40s & 378 & 178 & $41.80\%$ & $34.15\%$ & 622.60s & 405 & 206 & $49.70\%$ & $37.80\%$ & 656.70s & 446 & 227 & $63.10\%$ & $34.94\%$ \\
$8 \times 10^5$ & 739.20s & 402 & 192 & $38.00\%$ & $21.92\%$ & 781.90s & 473 & 229 & $45.20\%$ & $37.31\%$ & 816.30s & 508 & 262 & $53.30\%$ & $41.30\%$ & 862.70s & 556 & 287 & $67.60\%$ & $37.76\%$ \\

\hline
  \end{tabular}
 }
  \caption{Evaluation of {\sc DR-Arg} on various knowledge base sizes $|\KB|$ and fractions of conflicts $c$. The results represent averages from five runs per scenario.}
  \label{tab1}
\end{table*}

\section{Empirical Evaluations}

We present two forms of empirical evaluations -- a computational experiment and a human-user study. 

\subsection{Computational Experiments}
\label{comp-eval}

For our computational evaluation of DR-Arg, we utilize the following metrics to assess its performance:
\squishlist
\item \textbf{Dialogue Length $L$:} The total number of dialogue moves exchanged between the explainer and explainee agents.
\item \textbf{Dialogue Time $T$:} The duration of the dialogue, defined as the computational efforts required to generate arguments, assuming that communication cost is $0$.

\item \textbf{Number of Updates $N$:} The total count of updates to the explainee's knowledge base after the dialogue, reflecting the volume of new information incorporated.

\item \textbf{Change in Similarity $\Delta \Sigma$:} The change in the similarity between $KB_E$ and $KB_R$ (for $a =0.5$), comparing their initial (pre-interaction) and final (post-interaction) levels. 

\squishend

\subsubsection{Setup:} 
We created 16 unique pairs of $\KB_R$ and $\KB_E$ with sizes of $10^2 - 10^5$ by doing the following. (1)~We generated random inconsistent propositional KBs of varying sizes of $10^2 - 10^5$. (2)~We constructed $\KB_R$ by removing a minimal correction set (MCS) from the inconsistent KB to make them consistent.\footnote{A MCS is a $\subseteq$-minimal set of formulae whose removal renders an inconsistent KB consistent~\citep{marques-silva-ijcai13}.} (3)~To create $\KB_E$, we controlled the fraction of conflicts between the explainer and explainee with $c = |\KB_E| / |\KB_R|$. Specifically, starting with an empty $\KB_E$, we added formulae from MCS and, if needed, negations of random formulae from $\KB_R$ to meet the desired ratio. This process generated distinct KBs with conflict levels determined by $c$. (4)~Lastly, to have KBs of approximately the same size and with some similarity between them, we added a $1-c$ fraction of formulae from $\KB_R$ to $\KB_E$, as long as $\KB_E$ remained satisfiable. 

For generating arguments and counterarguments, we used a standard method from the literature~\citep{besnard2010mus}. The dialogue topic comprised a single \texttt{query} $\phi$, created by finding a formula entailed by $\KB_R$ but not by $\KB_E$. We identified this formula by examining the logical consequences of both knowledge bases. This process ensured the \texttt{query} addressed the knowledge discrepancy between the explainer and explainee, allowing to simulate a dialectical reconciliation dialogue.

We implemented a prototype of DR-Arg in Python using PySAT~\citep{imms-sat18}, and ran experiments with a time limit of $900s$ on a MacBook Pro machine with an M1 Max processor and 32GB of memory.\footnote{Code repository: \url{https://github.com/YODA-Lab/Dialectical-Reconciliation-with-Structured-Argumentation}.}


\subsubsection{Results:} Table \ref{tab1} presents the evaluation results of DR-Arg on various knowledge base sizes $|\KB|$ and fractions of conflicts $c$, allowing us to observe how they influence the dialogue time $T$, dialogue length $L$, number of updates $N$, the change in similarity with DR-Arg $\Delta \Sigma_{\text{\emph{DR}}}$, and the change in similarity with a state-of-the-art single-shot reconciliation approach $\Delta \Sigma_{\text{\emph{SSR}}}$~\citep{vas21}. 
The results reveal several trends and insights:

\squishlist

\item  Increasing $|\KB|$ led to longer dialogue times ($T$), reflecting the higher computational demand for larger knowledge bases.

\item Both the dialogue length ($L$) and the number of knowledge base updates ($N$) generally increased with larger $|\KB|$ and higher conflict ratios ($c$), indicating more extensive interactions required to resolve greater inconsistencies.

\item A noticeable increase in $\Delta \Sigma_{\text{\emph{DR}}}$ was observed with the rise in $N$, suggesting that more updates correlate with a greater improvement in the explainee’s understanding. Notably, $\Delta \Sigma_{\text{\emph{DR}}}$ consistently outperformed $\Delta \Sigma_{\text{\emph{SSR}}}$, underscoring the advantage of DR-Arg's iterative, multi-move approach over single-shot reconciliation methods.

\squishend

\subsection{Human User Study}
\label{sec:user-study}

We conducted a study involving the simulated scenario described in our motivating example (see Section~\ref{sec:motivating-example}). As a brief recap, a human user is presented with the task of troubleshooting an AI home assistant robot named ``Roomie'' that appears to be disconnected from the internet. The user is given a set of prompts to help them diagnose the problem, such as checking the associated mobile app, confirming Roomie's connection to the charging base, verifying Roomie's connection to the internet via a wired connector, and noting a flashing light next to the LAN port. However, the user is faced with several complications that hinder their ability to resolve the issue. These include an outdated mobile app, an expired license for the wired connection, and a low battery indicated by the flashing light. These obstacles create a realistic scenario for the user to navigate, as they must interact with Roomie to understand the underlying issues in order to get it up and running again.

Overall, this study provides a valuable opportunity to explore how humans interact with AI systems in real-world situations, and how they approach troubleshooting and problem-solving when faced with unexpected obstacles. From a technical standpoint, this narrative allowed us to approximate a human model, facilitating the use of a single-shot model reconciliation-based method as a baseline. A detailed setup of the study can be found in the appendix at the end of the document.

\subsubsection{Study Design:}
Participants were introduced to the problem through a narrative dialogue that explained the scenario's premise and known information. After posing the initial \texttt{query} ``Why are you disconnected?'', participants were divided into two groups:
\squishlist
    \item \textbf{Single-Shot (SSR)}: Group 1 received a single-shot model reconciliation explanation, where the human model was assumed to include the information provided during the scenario's introduction. The explanation was computed using the solver in~\citep{vas21}.

    \item \textbf{DR-Arg}: Group 2 interacted with DR-Arg's explanations, choosing from four unique questions (i.e., counterarguments) in a game-like format. They could continue asking questions or decide to end the interaction.

\squishend

Upon completing their interaction with Roomie, participants were asked four multiple-choice questions to evaluate their understanding of the issues, generating a comprehension score. They also responded to three Likert-scale questions (1: strongly disagree, 5: strongly agree) to gauge their satisfaction with the interaction and explanations, resulting in a satisfaction score. Our hypothesis was:

\begin{quote}
\textbf{H:} DR-Arg will achieve higher comprehension and satisfaction 
compared to the SSR.
\end{quote}

\subsubsection{Study Results and Discussion:}
We recruited 100 participants through Prolific \citep{palan2018prolific}, of whom 97 completed the study. The participants were diverse in terms of age, gender, and educational background, with all of them being proficient in English and having at least an undergraduate degree. They were compensated with a base payment of $\$2.50$ and had the opportunity to earn an additional $\$2.00$ bonus for correctly answering the comprehension questions.\footnote{The study was approved by our institution's ethics board and adhered to the guidelines for responsible research practices.}

In the DR-Arg group, participant engagement varied, leading us to further classify this group for analysis. Specifically, we divided the DR-Arg participants into two subgroups based on their interaction depth:
\squishlist
    \item \textbf{DR-Arg\textsubscript{Single}}: This subgroup is comprised of participants who chose to end the interaction after only one question. 
    \item \textbf{DR-Arg\textsubscript{Multi}}: This subgroup is comprised of participants who engaged with more than one question. 
\squishend
This classification allowed us to evaluate the impact of deeper interaction on comprehension and satisfaction.

The study results, presented in Table~\ref{tab:results}, display the average scores for comprehension and satisfaction, alongside the statistical significance of differences between the SSR and DR-Arg groups.

The results of the user study are presented in Table~\ref{tab:results}. As hypothesized, the DR-Arg participants outperformed the SSR group in terms of both comprehension and satisfaction scores. The differences between the two groups were statistically significant according to independent samples t-tests, with p-values below 0.05. 

The DR-Arg\textsubscript{Single} subgroup achieved better comprehension scores than the SSR group, suggesting that even a single interaction with DR-Arg can lead to improved understanding compared to a single-shot explanation. However, the most notable results were observed in the DR-Arg\textsubscript{Multi} subgroup, which obtained the highest comprehension and satisfaction scores among all groups. This finding highlights the effectiveness of deeper, multi-query interactions in dialectical reconciliation for enhancing user understanding and satisfaction.

\begin{table}[!t]
\centering
\renewcommand{\arraystretch}{1.2}
\resizebox{0.99\columnwidth}{!}{ 
\begin{tabular}{|c|c|c|c|c|}
\hline
\textbf{} & \textbf{SSR} & \textbf{DR-Arg} & \textbf{\begin{tabular}[c]{@{}c@{}}\textbf{DR-Arg\textsubscript{Single}}\end{tabular}} &\textbf{DR-Arg\textsubscript{Multi}}  \\ 
\hline
\hline
Number of Participants             & 49     & 48            & 11 & 37            \\ \hline
Comprehension Score (out of $4$)           & 0.30 & 2.60 & 1.18& \textbf{3.02}     \\ \hline
Satisfaction Score (out of $5$)            & 2.94  & 3.57  & 3.09 & \textbf{3.74} \\ \hline
\end{tabular}
}
\caption{Results of the user study.}
\label{tab:results}
\end{table}

As anticipated, the SSR participants scored lower on comprehension questions, possibly due to their inability to ask follow-up questions and only receiving information based on Roomie's assumed model of them. In contrast, the DR-Arg participants outperformed the SSR group, with the results being statistically significant according to a t-test with a p-value of 0.05. The DR-Arg\textsubscript{Single} subgroup showed improved comprehension over SSR, indicating that even minimal interaction with DR-Arg is more informative than a single-shot explanation. However, the most notable results were observed in the DR-Arg\textsubscript{Multi} subgroup, which achieved the highest comprehension and satisfaction scores.  This underscores the efficacy of deeper, multi-query interactions in dialectical reconciliation for enhancing understanding and user satisfaction. 

The study confirms our hypothesis \textbf{H}, illustrating that dialectical reconciliation is more effective in fostering understanding and addressing human user concerns than a single-shot approach. 


\section{Conclusions and Future Work}
\label{concl}

Argumentation is often advocated as suitable for explanation, but there are very few studies of its suitability to humans.~In this paper, we presented DR-Arg, a novel framework utilizing (structured deductive) argumentation for performing dialectical reconciliation between an explainer and an explainee. We conducted a thorough evaluation ``in the wild'', with our computational and human-user study results attesting to the efficacy of our approach.~These findings highlight the potential of argumentation-based approaches in enhancing the human-AI interaction of AI systems, particularly in domains where explainability is crucial.

Despite the promising aspects of our framework, it is important to acknowledge its limitations and potential areas for improvement. DR-Arg follows a fixed structure in presenting arguments and does not consider the effectiveness of personalizing the interactions according to the user's beliefs and preferences. DR-Arg also assumes seamless communication through well-defined dialogue moves, which may not reflect real-world complexities such as miscommunication or uncertainty. Finally, the current framework is limited to deductive argumentation and propositional logic, which may not be sufficient to express complex relationships and dependencies in real-world domains.

To address these limitations, we suggest the following future directions: (1)~Develop an adaptive approach that tailors arguments to individual users' needs and preferences based on user feedback and prior interactions \cite{sreedharan2021using,vasileioua2023please}. In \citet{tang2024approximating}, we have taken a preliminary step towards this end by proposing a probabilistic framework to approximate human user models from argumentation-based dialogues; (2)~Integrate DR-Arg with large language models \cite{brown2020language} to translate formal arguments and logical structures into intuitive, natural language expressions, enhancing accessibility and user-friendliness while maintaining logical coherence; and (3)~Consider alternative structured argumentation frameworks, such as ABA \cite{bondarenko1997abstract,vcyras2016aba+} or probabilistic argumentation frameworks \cite{kohlas2003probabilistic,hunter2013probabilistic}, to enable more complex reasoning and argument generation for a wider range of real-world problems.

\section*{Acknowledgements}

Stylianos Loukas Vasileiou, Ashwin Kumar, and William Yeoh are partially supported by the National Science Foundation (NSF) under award 2232055. Tran Cao Son is partially supported by NSF under awards 1914635 and 2151254 and by a subcontract from Wallaroo.AI. Francesca Toni is partially funded by the European Research Council (ERC) under the EU's Horizon 2020 research and innovation programme (grant agreement 101020934), by J.P. Morgan, and by the
Royal Academy of Engineering, UK, under the Research Chairs and Senior Research Fellowships scheme. The views and conclusions contained in this document are those of the authors and should not be interpreted as representing the official policies, either expressed or implied, of the sponsoring organizations, agencies, or governments.

\bibliographystyle{kr}
\bibliography{kr-sample}

\begin{thebibliography}{}

\bibitem[\protect\citeauthoryear{Besnard and Hunter}{2001}]{besnard2001logic}
Besnard, P., and Hunter, A.
\newblock 2001.
\newblock A logic-based theory of deductive arguments.
\newblock {\em Artificial Intelligence}  203--235.

\bibitem[\protect\citeauthoryear{Besnard and
  Hunter}{2014}]{besnard2014constructing}
Besnard, P., and Hunter, A.
\newblock 2014.
\newblock Constructing argument graphs with deductive arguments: A tutorial.
\newblock {\em Argument \& Computation} 5(1):5--30.

\bibitem[\protect\citeauthoryear{Besnard \bgroup et al\mbox.\egroup
  }{2010}]{besnard2010mus}
Besnard, P.; Gr{\'e}goire, {\'E}.; Piette, C.; and Raddaoui, B.
\newblock 2010.
\newblock {MUS}-based generation of arguments and counter-arguments.
\newblock In {\em Proceedings of IRI},  239--244.

\bibitem[\protect\citeauthoryear{Black and Hunter}{2009}]{black2009inquiry}
Black, E., and Hunter, A.
\newblock 2009.
\newblock An inquiry dialogue system.
\newblock {\em Autonomous Agents and Multi-Agent Systems} 19:173--209.

\bibitem[\protect\citeauthoryear{Black, Maudet, and
  Parsons}{2021}]{black2021argumentation}
Black, E.; Maudet, N.; and Parsons, S.
\newblock 2021.
\newblock Argumentation-based dialogue.
\newblock {\em Handbook of Formal Argumentation} 2.

\bibitem[\protect\citeauthoryear{Bondarenko \bgroup et al\mbox.\egroup
  }{1997}]{bondarenko1997abstract}
Bondarenko, A.; Dung, P.~M.; Kowalski, R.~A.; and Toni, F.
\newblock 1997.
\newblock An abstract, argumentation-theoretic approach to default reasoning.
\newblock {\em Artificial intelligence} 93(1-2):63--101.

\bibitem[\protect\citeauthoryear{Brown \bgroup et al\mbox.\egroup
  }{2020}]{brown2020language}
Brown, T.; Mann, B.; Ryder, N.; Subbiah, M.; Kaplan, J.~D.; Dhariwal, P.;
  Neelakantan, A.; Shyam, P.; Sastry, G.; Askell, A.; et~al.
\newblock 2020.
\newblock Language models are few-shot learners.
\newblock In {\em Proceedings of NeurIPS},  1877--1901.

\bibitem[\protect\citeauthoryear{Bud{\'a}n \bgroup et al\mbox.\egroup
  }{2020}]{budan2020proximity}
Bud{\'a}n, M.~C.; Cobo, M.~L.; Martinez, D.~C.; and Simari, G.~R.
\newblock 2020.
\newblock Proximity semantics for topic-based abstract argumentation.
\newblock {\em Information Sciences} 508:135--153.

\bibitem[\protect\citeauthoryear{Chakraborti \bgroup et al\mbox.\egroup
  }{2017}]{chakraborti2017plan}
Chakraborti, T.; Sreedharan, S.; Zhang, Y.; and Kambhampati, S.
\newblock 2017.
\newblock Plan explanations as model reconciliation: Moving beyond explanation
  as soliloquy.
\newblock In {\em Proceedings of IJCAI},  156--163.

\bibitem[\protect\citeauthoryear{Collins, Magazzeni, and
  Parsons}{2019}]{collins2019towards}
Collins, A.; Magazzeni, D.; and Parsons, S.
\newblock 2019.
\newblock Towards an argumentation-based approach to explainable planning.
\newblock In {\em Proceedings of XAIP},  39--43.

\bibitem[\protect\citeauthoryear{{\v{C}}yras and Toni}{2016}]{vcyras2016aba+}
{\v{C}}yras, K., and Toni, F.
\newblock 2016.
\newblock Aba+ assumption-based argumentation with preferences.
\newblock In {\em Proceedings of KR},  553--556.

\bibitem[\protect\citeauthoryear{{\v{C}}yras \bgroup et al\mbox.\egroup
  }{2021}]{ijcai2021p0600}
{\v{C}}yras, K.; Rago, A.; Albini, E.; Baroni, P.; and Toni, F.
\newblock 2021.
\newblock Argumentative {XAI:} {A} survey.
\newblock In {\em Proceedings of IJCAI},  4392--4399.

\bibitem[\protect\citeauthoryear{Dennis and Oren}{2022}]{dennis2022explaining}
Dennis, L.~A., and Oren, N.
\newblock 2022.
\newblock Explaining {BDI} agent behaviour through dialogue.
\newblock {\em Autonomous Agents and Multi-Agent Systems} 36(2):29.

\bibitem[\protect\citeauthoryear{Dice}{1945}]{dice1945measures}
Dice, L.~R.
\newblock 1945.
\newblock Measures of the amount of ecologic association between species.
\newblock {\em Ecology} 26(3):297--302.

\bibitem[\protect\citeauthoryear{Dung and Son}{2022}]{DungS22a}
Dung, H.~T., and Son, T.~C.
\newblock 2022.
\newblock On model reconciliation: How to reconcile when robot does not know
  human's model?
\newblock In {\em Proceedings of ICLP}, volume 364,  27--48.

\bibitem[\protect\citeauthoryear{Fan and Toni}{2012}]{fan2012agent}
Fan, X., and Toni, F.
\newblock 2012.
\newblock Agent strategies for {ABA}-based information-seeking and inquiry
  dialogues.
\newblock In {\em Proceedings of ECAI},  324--329.

\bibitem[\protect\citeauthoryear{Fan and Toni}{2015}]{FanT15}
Fan, X., and Toni, F.
\newblock 2015.
\newblock On computing explanations in argumentation.
\newblock In {\em Proceedings of AAAI},  1496--1502.

\bibitem[\protect\citeauthoryear{Gordon}{1994}]{gordon1994}
Gordon, T.~F.
\newblock 1994.
\newblock An inquiry dialogue system.
\newblock {\em Artificial Intelligence and Law} 2:239--292.

\bibitem[\protect\citeauthoryear{Hamblin}{1970}]{hamblin1970fallacies}
Hamblin, C.~L.
\newblock 1970.
\newblock {\em Fallacies}.
\newblock Methuen and Co. Ltd.

\bibitem[\protect\citeauthoryear{Hamblin}{1971}]{hamblin1971mathematical}
Hamblin, C.~L.
\newblock 1971.
\newblock Mathematical models of dialogue.
\newblock {\em Theoria} 37(2):130--155.

\bibitem[\protect\citeauthoryear{Hitchcock and
  Hitchcock}{2017}]{hitchcock2017some}
Hitchcock, D., and Hitchcock, D.
\newblock 2017.
\newblock Some principles of rational mutual inquiry.
\newblock {\em On Reasoning and Argument: Essays in Informal Logic and on
  Critical Thinking}  313--321.

\bibitem[\protect\citeauthoryear{Hunter}{2013}]{hunter2013probabilistic}
Hunter, A.
\newblock 2013.
\newblock A probabilistic approach to modelling uncertain logical arguments.
\newblock {\em International Journal of Approximate Reasoning} 54(1):47--81.

\bibitem[\protect\citeauthoryear{Ignatiev, Morgado, and
  Marques{-}Silva}{2018}]{imms-sat18}
Ignatiev, A.; Morgado, A.; and Marques{-}Silva, J.
\newblock 2018.
\newblock {PySAT:} {A} {Python} toolkit for prototyping with {SAT} oracles.
\newblock In {\em Proceedings of SAT},  428--437.

\bibitem[\protect\citeauthoryear{Jackson}{1992}]{jackson1992computing}
Jackson, P.
\newblock 1992.
\newblock Computing prime implicates.
\newblock In {\em Proceedings of CSC},  65--72.

\bibitem[\protect\citeauthoryear{Johnson-Laird}{1983}]{johnson1983mental}
Johnson-Laird, P.~N.
\newblock 1983.
\newblock {\em Mental Models: Towards a Cognitive Science of Language,
  Inference, and Consciousness}.

\bibitem[\protect\citeauthoryear{Kambhampati}{2019}]{kambhampati2019synthesizing}
Kambhampati, S.
\newblock 2019.
\newblock Synthesizing explainable behavior for human-{AI} collaboration.
\newblock In {\em Proceedings of AAMAS},  1--2.

\bibitem[\protect\citeauthoryear{Kohlas}{2003}]{kohlas2003probabilistic}
Kohlas, J.
\newblock 2003.
\newblock Probabilistic argumentation systems: A new way to combine logic with
  probability.
\newblock {\em Journal of Applied Logic} 1(3-4):225--253.

\bibitem[\protect\citeauthoryear{Marques{-}Silva \bgroup et al\mbox.\egroup
  }{2013}]{marques-silva-ijcai13}
Marques{-}Silva, J.; Heras, F.; Janota, M.; Previti, A.; and Belov, A.
\newblock 2013.
\newblock On computing minimal correction subsets.
\newblock In {\em Proceedings of IJCAI},  615--622.

\bibitem[\protect\citeauthoryear{Mercier and Sperber}{2011}]{mercier2011humans}
Mercier, H., and Sperber, D.
\newblock 2011.
\newblock Why do humans reason? {Arguments} for an argumentative theory.
\newblock {\em Behavioral and Brain Sciences} 34(2):57--74.

\bibitem[\protect\citeauthoryear{Oren, van Deemter, and
  Vasconcelos}{2020}]{oren2020argument}
Oren, N.; van Deemter, K.; and Vasconcelos, W.~W.
\newblock 2020.
\newblock Argument-based plan explanation.
\newblock In {\em Knowledge Engineering Tools and Techniques for {AI}
  Planning}.
\newblock  173--188.

\bibitem[\protect\citeauthoryear{Palan and Schitter}{2018}]{palan2018prolific}
Palan, S., and Schitter, C.
\newblock 2018.
\newblock Prolific. ac—a subject pool for online experiments.
\newblock {\em Journal of Behavioral and Experimental Finance} 17:22--27.

\bibitem[\protect\citeauthoryear{Parkes}{1997}]{parkes1997clustering}
Parkes, A.~J.
\newblock 1997.
\newblock Clustering at the phase transition.
\newblock In {\em Proceedings of AAAI},  340--345.

\bibitem[\protect\citeauthoryear{Parsons, Wooldridge, and
  Amgoud}{2002}]{parsons2002analysis}
Parsons, S.; Wooldridge, M.; and Amgoud, L.
\newblock 2002.
\newblock An analysis of formal inter-agent dialogues.
\newblock In {\em Proceedings of AAMAS},  394--401.

\bibitem[\protect\citeauthoryear{Parsons, Wooldridge, and
  Amgoud}{2003}]{parsons2003properties}
Parsons, S.; Wooldridge, M.; and Amgoud, L.
\newblock 2003.
\newblock Properties and complexity of some formal inter-agent dialogues.
\newblock {\em Journal of Logic and Computation} 13(3):347--376.

\bibitem[\protect\citeauthoryear{Plotkin}{1981}]{plotkin1981structural}
Plotkin, G.~D.
\newblock 1981.
\newblock {\em A Structural Approach to Operational Semantics}.
\newblock Aarhus University.

\bibitem[\protect\citeauthoryear{Prakken}{2006}]{prakken2006formal}
Prakken, H.
\newblock 2006.
\newblock Formal systems for persuasion dialogue.
\newblock {\em The Knowledge Engineering Review} 21(2):163--188.

\bibitem[\protect\citeauthoryear{Rago, Li, and
  Toni}{2023}]{rago2023interactive}
Rago, A.; Li, H.; and Toni, F.
\newblock 2023.
\newblock Interactive explanations by conflict resolution via argumentative
  exchanges.
\newblock In {\em Proceedings of KR},  582--592.

\bibitem[\protect\citeauthoryear{Son \bgroup et al\mbox.\egroup
  }{2021}]{son2021model}
Son, T.~C.; Nguyen, V.; Vasileiou, S.~L.; and Yeoh, W.
\newblock 2021.
\newblock Model reconciliation in logic programs.
\newblock In {\em Proceedings of ECAI},  393--406.

\bibitem[\protect\citeauthoryear{Sorensen}{1948}]{sorensen1948method}
Sorensen, T.~A.
\newblock 1948.
\newblock A method of establishing groups of equal amplitude in plant sociology
  based on similarity of species content and its application to analyses of the
  vegetation on danish commons.
\newblock {\em Kongelige Danske Videnskabernes Selskab} 5:1--34.

\bibitem[\protect\citeauthoryear{Sreedharan, Chakraborti, and
  Kambhampati}{2021}]{sreedharan2021foundations}
Sreedharan, S.; Chakraborti, T.; and Kambhampati, S.
\newblock 2021.
\newblock Foundations of explanations as model reconciliation.
\newblock {\em Artificial Intelligence} 301:103558.

\bibitem[\protect\citeauthoryear{Sreedharan, Kulkarni, and
  Kambhampati}{2022}]{sreedharan2022explainable}
Sreedharan, S.; Kulkarni, A.; and Kambhampati, S.
\newblock 2022.
\newblock Explainable human--{AI} interaction: A planning perspective.
\newblock {\em Synthesis Lectures on Artificial Intelligence and Machine
  Learning} 16(1):1--184.

\bibitem[\protect\citeauthoryear{Sreedharan, Srivastava, and
  Kambhampati}{2021}]{sreedharan2021using}
Sreedharan, S.; Srivastava, S.; and Kambhampati, S.
\newblock 2021.
\newblock Using state abstractions to compute personalized contrastive
  explanations for {AI} agent behavior.
\newblock {\em Artificial Intelligence} 301:103570.

\bibitem[\protect\citeauthoryear{Tang, Vasileiou, and
  Yeoh}{2024}]{tang2024approximating}
Tang, Y.; Vasileiou, S.~L.; and Yeoh, W.
\newblock 2024.
\newblock Approximating human models during argumentation-based dialogues.
\newblock {\em arXiv preprint arXiv:2405.18650}.

\bibitem[\protect\citeauthoryear{Teze, Godo, and
  Simari}{2022}]{teze2022approach}
Teze, J.~C.; Godo, L.; and Simari, G.~I.
\newblock 2022.
\newblock An approach to improve argumentation-based epistemic planning with
  contextual preferences.
\newblock {\em International Journal of Approximate Reasoning} 151:130--163.

\bibitem[\protect\citeauthoryear{Vasileiou and
  Yeoh}{2023}]{vasileioua2023please}
Vasileiou, S.~L., and Yeoh, W.
\newblock 2023.
\newblock {PLEASE}: Generating personalized explanations in human-aware
  planning.
\newblock In {\em Proceedings of ECAI},  2411--2418.

\bibitem[\protect\citeauthoryear{Vasileiou \bgroup et al\mbox.\egroup
  }{2022}]{vasileioulogic}
Vasileiou, S.~L.; Yeoh, W.; Son, T.~C.; Kumar, A.; Cashmore, M.; and Magazzeni,
  D.
\newblock 2022.
\newblock A logic-based explanation generation framework for classical and
  hybrid planning problems.
\newblock {\em Journal of Artificial Intelligence Research} 73:1473--1534.

\bibitem[\protect\citeauthoryear{Vasileiou, Previti, and Yeoh}{2021}]{vas21}
Vasileiou, S.~L.; Previti, A.; and Yeoh, W.
\newblock 2021.
\newblock On exploiting hitting sets for model reconciliation.
\newblock In {\em Proceedings of AAAI},  6514--6521.

\bibitem[\protect\citeauthoryear{Walton and
  Krabbe}{1995}]{walton1995commitment}
Walton, D., and Krabbe, E.~C.
\newblock 1995.
\newblock {\em Commitment in Dialogue: Basic Concepts of Interpersonal
  Reasoning}.
\newblock SUNY press.

\end{thebibliography}

\newpage

\section*{Appendix}

\subsection*{Human-User Study}

\noindent \textbf{Demographics:} Overall, 97 participants completed the study. All participants were proficient in English and had at leas an undergraduate education. Out of the 97 participants, 59 identified as female, 35 as male, and 3 as other.

\subsubsection*{Study Details} 

All participants first received the information depicted in Figure~1.~Afterwards, the participants answered two attention check questions, they were divided into two groups:
\squishlist
    \item \textbf{Single-Shot (SSR)}: Group 1 received a single-shot model reconciliation explanation, where the human model was assumed to include the information provided during the scenario's introduction. The explanations were computed using a state-of-the-art solver by Vasileiou et al. [2021].

    \item \textbf{DR-Arg}: Group 2 interacted with DR-A's explanations, choosing from four unique questions (counterarguments to Roomie's responses) in a game-like format. They could continue asking questions or decide to end the interaction.

\squishend

Figures~2 to 7 show some of the interactions the DR-Arg users had with Roomie, while Figure~8 shows the interaction the SSR users had with Roomie.

\subsection*{Study Questions and More Results} 

In the DR-Arg group, participant engagement varied, leading us to further classify this group for analysis. Specifically, we divided the DR-Arg participants into two subgroups based on their interaction depth:
\squishlist
    \item \textbf{DR-Arg\textsubscript{Single}}: This subgroup is comprised of participants who chose to end the interaction after only one question. 
    \item \textbf{DR-Arg\textsubscript{Multi}}: This subgroup is comprised of participants who engaged with more than one question. 
\squishend
This classification allowed us to evaluate the impact of deeper interaction on comprehension and satisfaction.

After the participants in each group interacted with Roomie, they were all asked to answer the questions below. The answers to these questions can be seen in Figures~9~and~10. \\

\noindent \textbf{Comprehension questions:}
\begin{itemize}
\item[Q1.] \textit{Why did Roomie not have an internet connection?}
\begin{enumerate}

    \item Hardware lock. (\textbf{Correct answer})
    \item Cable not connected properly.
    \item Wifi was not working.
    \item Docking station failure.
\end{enumerate}
\item[Q2.] \textit{Why were there issues even though you paid for the full package?}
\begin{enumerate}
    \item License expired. (\textbf{Correct answer})
    \item Service required.
    \item Roomie was set up incorrectly.
    \item Roomie does not support uneven floors.
\end{enumerate}
\item[Q3.] \textit{Why was there a flashing light next to the internet port?}
\begin{enumerate}
    \item Battery was low. (\textbf{Correct answer})
    \item Internet port was in use.
    \item Roomie was malfunctioning.
    \item Roomie's connection was high-speed.
\end{enumerate}

\item[Q4.] \textit{Why did the app say Roomie was connected to the internet?}
\begin{enumerate}
    \item Roomie was connected to the internet.
    \item App was outdated. (\textbf{Correct answer})
    \item All cables were securely connected.
    \item Roomie was malfunctioning.
\end{enumerate}
\end{itemize}

\vspace{5mm}
\noindent \textbf{Likert-type questions:}

\begin{itemize}
\item[Q1.] \textit{Roomie's explanations were easy to understand.} 
\begin{itemize}
    \item[] 1: \emph{Strongly disagree} - 5 : \emph{Strongly agree}
\end{itemize}
\item[Q2.] \textit{I understood all the issues with Roomie.} \begin{itemize}
    \item[]  1: \emph{Strongly disagree} - 5 : \emph{Strongly agree}
\end{itemize}
\item[Q3.] \textit{I would have liked to ask more questions to improve my understanding.} \begin{itemize}
    \item[] 1: \emph{Strongly disagree} - 5 : \emph{Strongly agree}
\end{itemize}
\end{itemize}

\vspace{100pt}
\begin{figure}[!ht]
     \centering
     \includegraphics[width=1.\columnwidth]{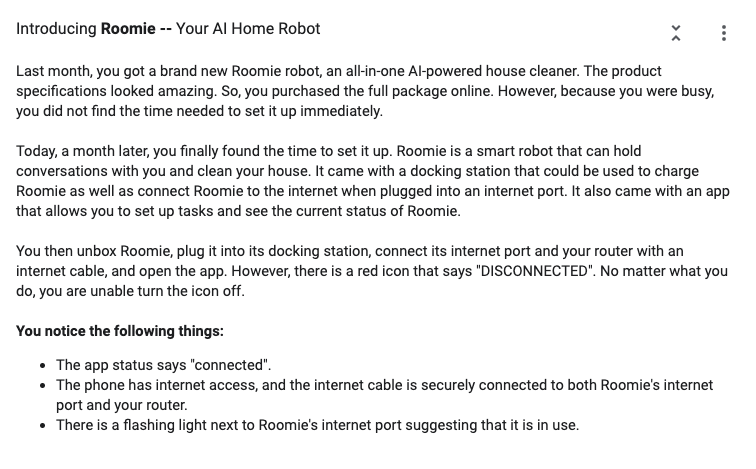}
     \caption{The introductory information shown to participants at the beginning of the study.}
 \end{figure}

\begin{figure}[!h]
    \centering
        \centering
        \includegraphics[width=.8\columnwidth]{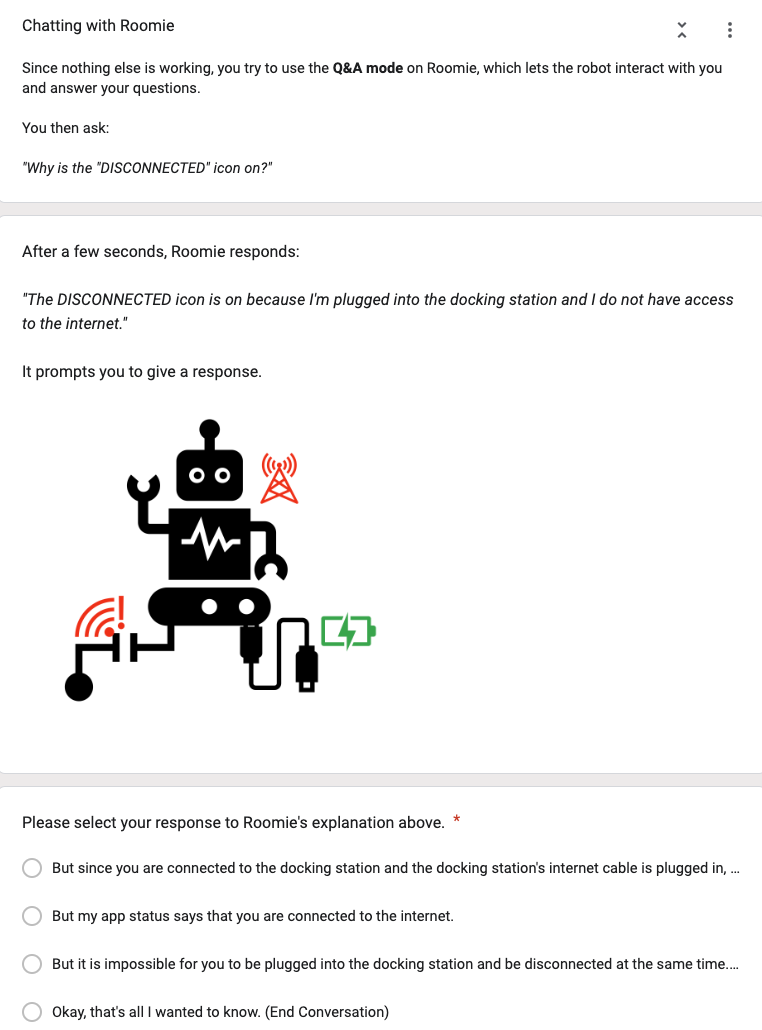} 
        \caption{(DR-Arg) Initial interaction: starting query, response (support) to query, and follow-up questions (refute).}
\end{figure}

\begin{figure}[!ht]
    \centering
        \centering
        \includegraphics[width=.8\columnwidth]{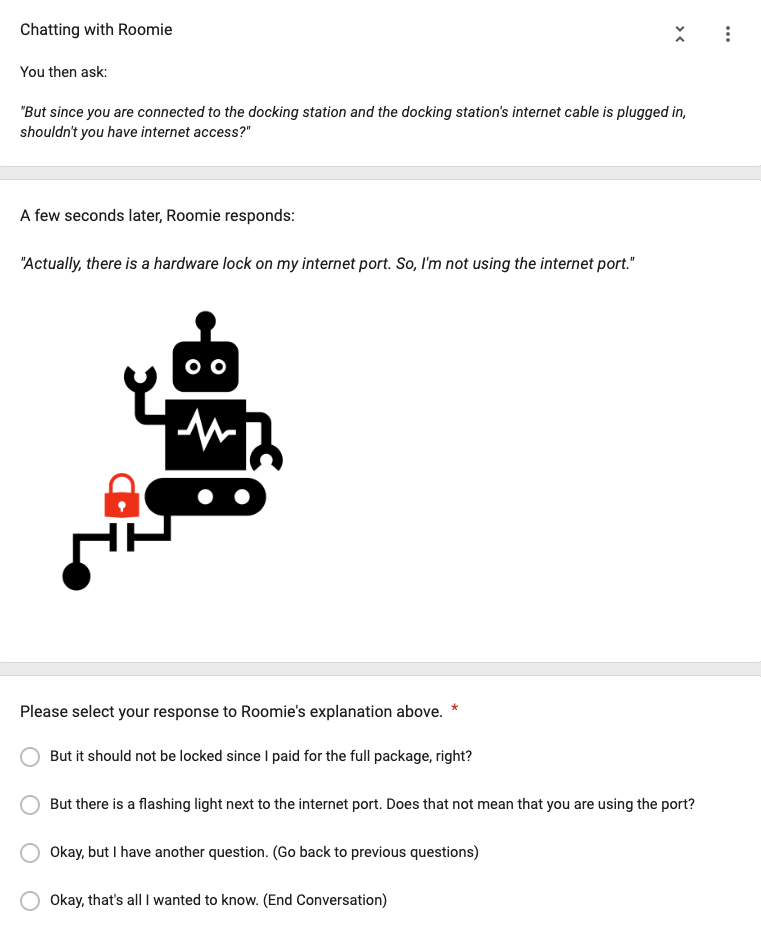}
        \caption{(DR-Arg) Second interaction: Refutation to user response, and follow-up questions.}
\end{figure}

\begin{figure}[!ht]
    \centering
        \centering
        \includegraphics[width=.8\columnwidth]{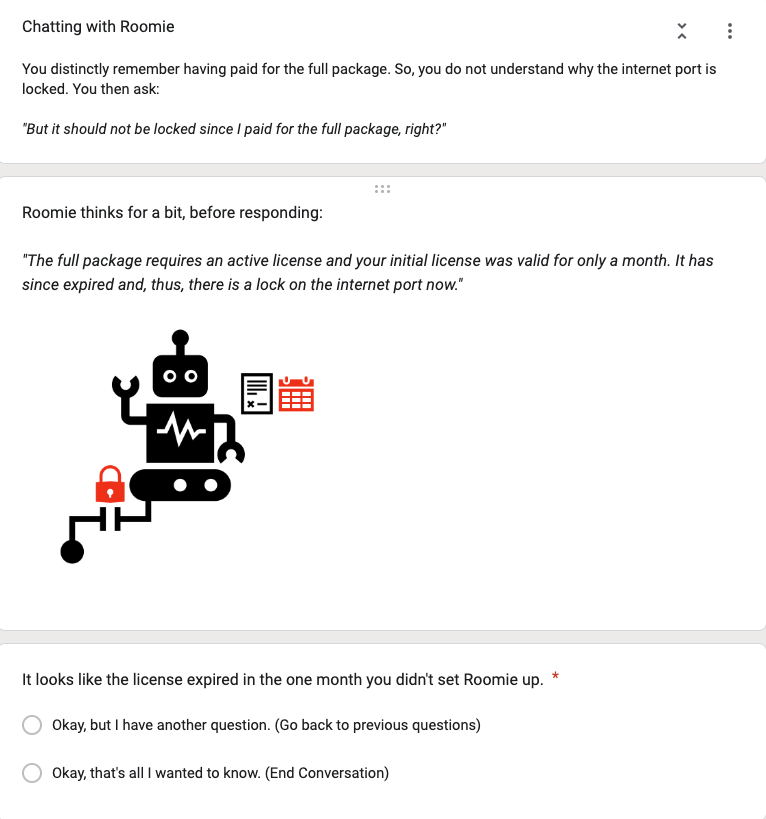}
        \caption{(DR-Arg) Third interaction: Refutation to user response, and follow-up questions.}
\end{figure}

\begin{figure}[!ht]
    \centering
        \centering
        \includegraphics[width=.8\columnwidth]{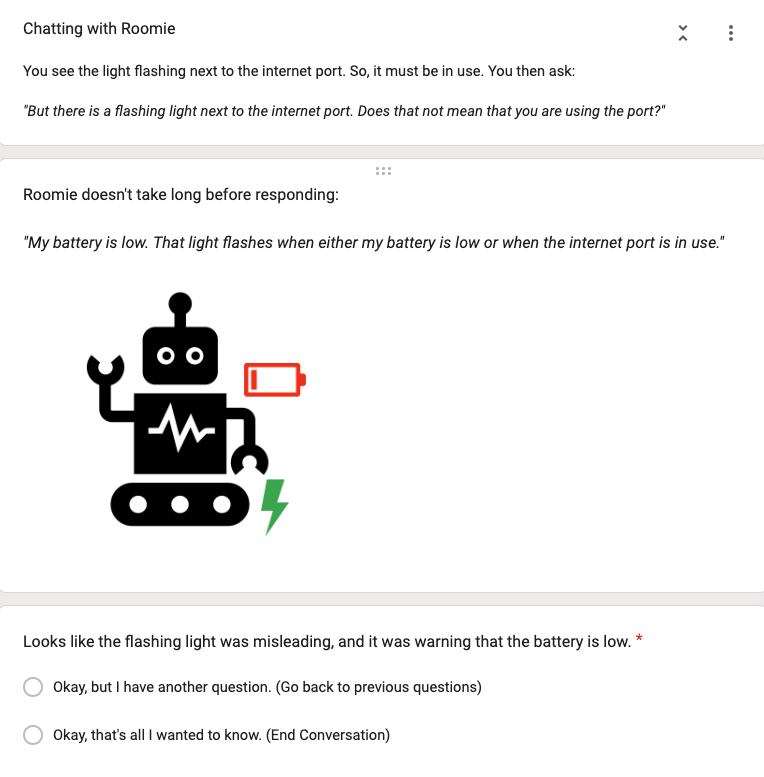}
        \caption{(DR-Arg) Fourth interaction: Refutation to user response, and follow-up questions.}
\end{figure}

\begin{figure}[!ht]
    \centering
        \centering
        \includegraphics[width=.8\columnwidth]{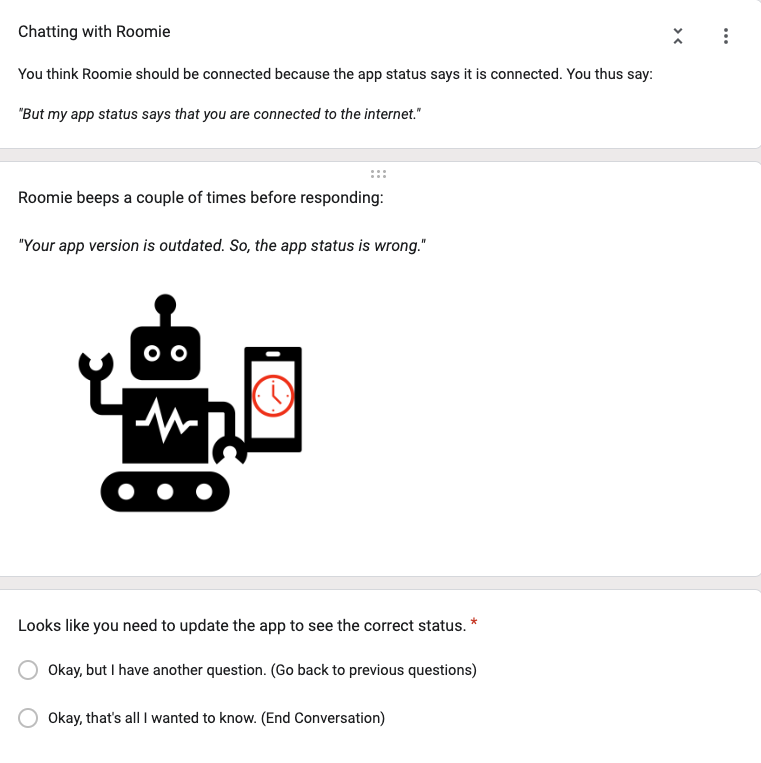}
        \caption{(DR-Arg) Fifth interaction: Refutation to user response, and follow-up questions.}
\end{figure}

\begin{figure}[!ht]
    \centering
        \centering
        \includegraphics[width=.8\columnwidth]{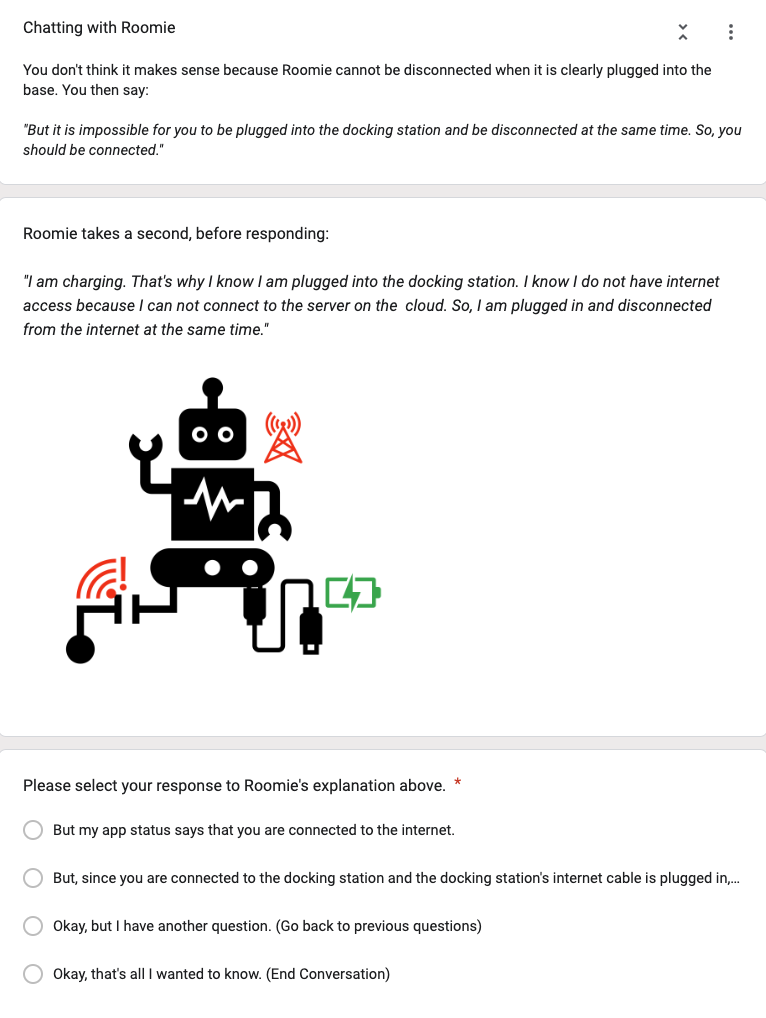}
        \caption{(DR-Arg) Sixth interaction: Refutation to user response, and follow-up questions.}
\end{figure}

\begin{figure}[!ht]
\centering
  \includegraphics[width=.8\columnwidth]{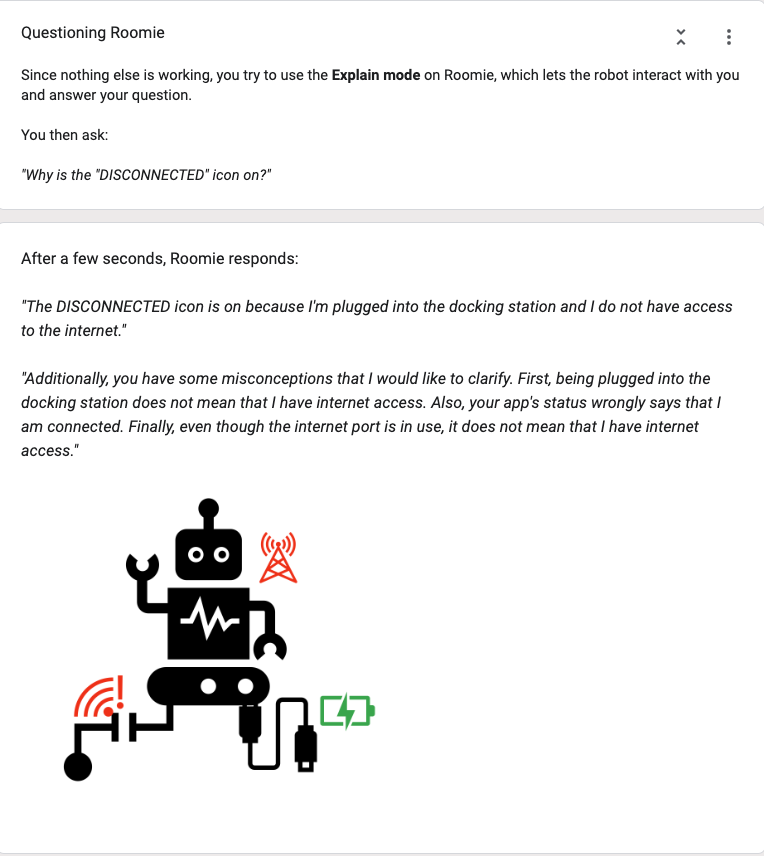}
  \caption{(SSR) Initial query and response.}
\end{figure}

\begin{figure}[!ht]
\centering
  \includegraphics[width=0.9\columnwidth]{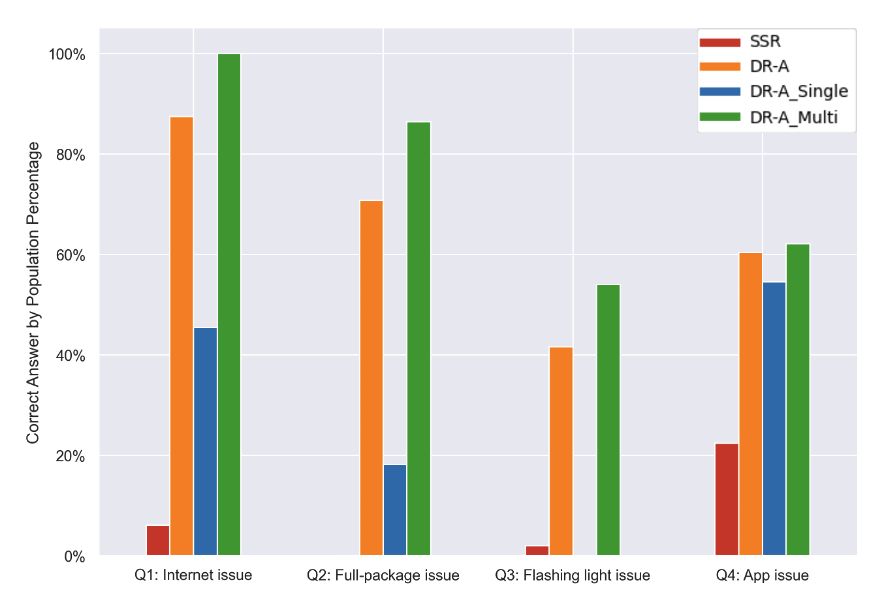}
  \caption{Distribution of answers to comprehension questions.}
 \end{figure}

\begin{figure}[!ht]
\centering
   \includegraphics[width=.9\columnwidth]{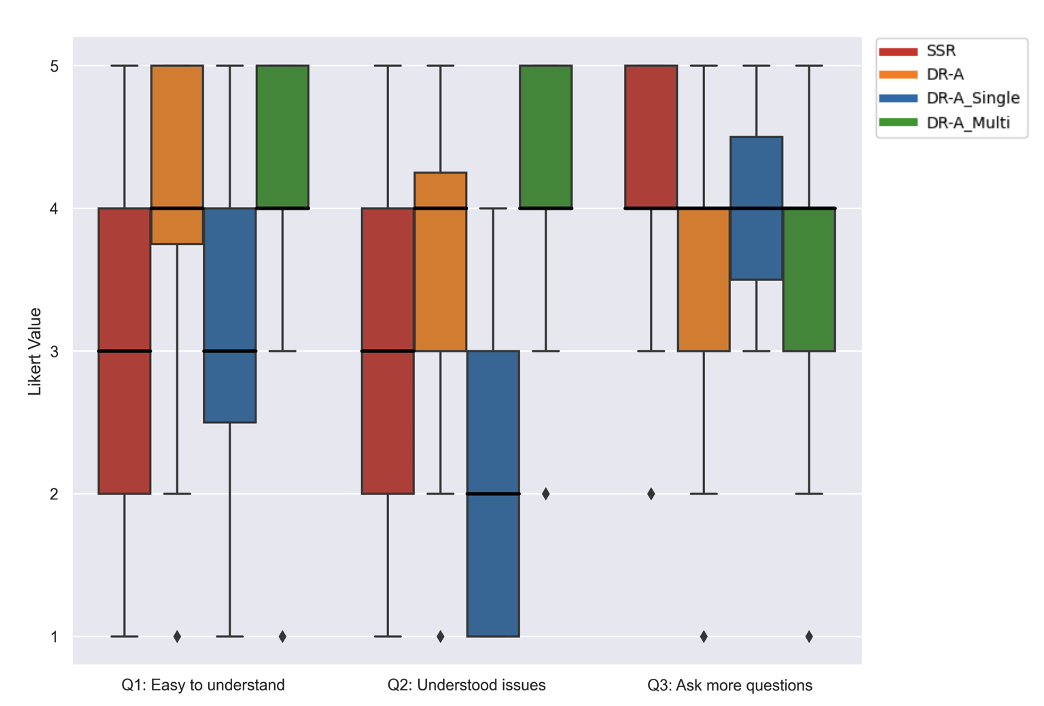}
  \caption{Distribution of answers to Likert-type questions.}
\end{figure}

\end{document}